\documentclass[journal,comsoc]{IEEEtran}
\IEEEoverridecommandlockouts
\usepackage{cite}
\usepackage{amsmath,amssymb,amsfonts}
\usepackage{algorithmicx}
\usepackage{algorithm}
\usepackage{algpseudocode}
\usepackage{graphicx}
\usepackage{textcomp}
\usepackage{float}
\usepackage{subfigure}
\usepackage{amsmath,amssymb}
\usepackage{bm}
\usepackage{bbm}
\usepackage{multirow} 
\usepackage{caption}
\usepackage{booktabs}
\usepackage{array}
\usepackage[table]{xcolor}
\usepackage{hyperref}

\newcommand{\matr}[1]{\mathbf{#1}}

\newcommand{\set}[1]{\mathcal{#1}}

\newcommand{\Rev}[1]{\textcolor{black}{#1}}

\usepackage{xcolor}

\def\BibTeX{{\rm B\kern-.05em{\sc i\kern-.025em b}\kern-.08em
		T\kern-.1667em\lower.7ex\hbox{E}\kern-.125emX}}
\usepackage[T1]{fontenc}% optional T1 font encoding

\usepackage{comment}

\begin{document}

\title{Exploring Diversity-based Active Learning for 3D Object Detection in Autonomous Driving}

\author{Zhihao~Liang,~\IEEEmembership{Student Member,~IEEE,}}% <-this % stops a space
\author{
        Jinpeng~Lin,
        Zhihao~Liang,
        Shengheng~Deng,
        Lile~Cai,
        Tao~Jiang,
        Tianrui~Li,
        Kui~Jia, % <-this % stops a space
        Xun~Xu,~\IEEEmembership{Senior Member,~IEEE,}
\thanks{Jinpeng~Lin and Zhihao~Liang contributed equally to this work. Correspondence to: Xun~Xu e-mail: alex.xun.xu@gmail.com .}
\thanks{Xun~Xu and Lile~Cai are with I2R, A-STAR, Singapore.}% <-this % stops a space
\thanks{Zhihao~Liang, Shengheng~Deng and Kui~Jia are with South China University of Technology. }%
\thanks{Jinpeng~Lin and Tianrui~Li are with Southwest Jiaotong University.}
\thanks{Tao~Jiang is with Chengdu University of Information Technology.}
%\thanks{This research is supported by the National Natural Science Foundation of China (NSFC) under Grant 62106078, Sichuan Science and Technology Program (Project No.: 2023NSFSC1421), and the Agency for Science, Technology and Research, Singapore (A*STAR) (Grant No. C210112059 and M23L7b0021).}
}

%{Shell \MakeLowercase{\textit{et al.}}: Bare Demo of IEEEtran.cls for IEEE Communications Society Journals}
\maketitle
\begin{abstract}
3D object detection has recently received much attention due to its great potential in autonomous vehicle (AV). The success of deep learning based object detectors relies on the availability of large-scale annotated datasets, which is time-consuming and expensive to compile, especially for 3D bounding box annotation.  In this work, we investigate diversity-based active learning (AL) as a potential solution to alleviate the annotation burden. Given limited annotation budget, only the most informative frames and objects are automatically selected for human to annotate. Technically, we take the advantage of the multimodal information provided in an AV dataset, and propose a novel acquisition function that enforces spatial and temporal diversity in the selected samples. We benchmark the proposed method against other AL strategies under realistic annotation cost measurements, where the realistic costs for annotating a frame and a 3D bounding box are both taken into consideration. We demonstrate the effectiveness of the proposed method on the nuScenes dataset and show that it outperforms existing AL strategies significantly. Code is available at \href{https://github.com/Linkon87/Exploring-Diversity-based-Active-Learning-for-3D-Object-Detection-in-Autonomous-Driving}{\textcolor{blue}{link}}.
\end{abstract}


\IEEEpeerreviewmaketitle

\section{Introduction}
%%% why active learning %%%
3D object detection has recently received much attention from the 3D computer vision community due to its great potential in autonomous vehicles (AVs). Most research has focused on enhancing detection accuracy through the design of more effective network architectures \cite{yang20203dssd, zhu2019class, yin2021center} and improved input representations \cite{qi2018frustum, Shi_2019_CVPR, Lang_2019_CVPR, yan2018second}. {The success of these deep learning-based methods heavily depends on large-scale annotated datasets, which are both time-consuming and costly to produce. Consequently, developing data-efficient learning techniques for 3D object detection is crucial to minimizing the effort required for annotation.}

% Most works are focused on improving detection accuracy by designing more effective network architectures \cite{yang20203dssd, zhu2019class, yin2021center} and input representations \cite{qi2018frustum, Shi_2019_CVPR, Lang_2019_CVPR, yan2018second}. The success of these deep learning-based methods relies on the availability of large-scale annotated datasets, the collection of which is time-consuming and prohibitively expensive. This is particularly true for 3D bounding box annotation in point cloud data, where the annotator needs to adjust the yaw angle frequently for accurate labeling. Therefore, the development of data-efficient learning techniques for 3D object detection is necessary to reduce the labeling efforts.

{Active learning (AL) presents a promising approach to alleviating the annotation burden. With a constrained annotation budget, AL seeks to maximize model performance by iteratively selecting the most informative samples for labeling, based on the model's current state. Depending on how informativeness is assessed, various methods can be broadly categorized into uncertainty-based \cite{roth2006margin, joshi2009multi, yoo2019learning}, diversity-based \cite{sener2017active, lin2017active}, and hybrid approaches \cite{yang2015multi, elhamifar2013convex, guo2010active}. }

% Active learning (AL) offers a promising solution to mitigate the annotation burden. Given a limited annotation budget, AL aims to maximize model performance by iteratively selecting the most informative samples to label based on the current model state. Depending on how informativeness is measured, various methods can be generally categorized into uncertainty-based \cite{roth2006margin, joshi2009multi, yoo2019learning}, diversity-based \cite{sener2017active, lin2017active}, and hybrid methods \cite{yang2015multi, elhamifar2013convex, guo2010active}. 
% Active learning has been successfully demonstrated in various computer vision tasks including image classification \cite{gal2015bayesian, gal2017deep, li2017dropout}, 2D object detection \cite{aghdam2019active,roy2018deep,kao2018localization}, and semantic segmentation \cite{mackowiak2018cereals, cai2021exploring, cai2021revisiting}.

\begin{figure*}
        \centering
        \includegraphics[width=1\linewidth]{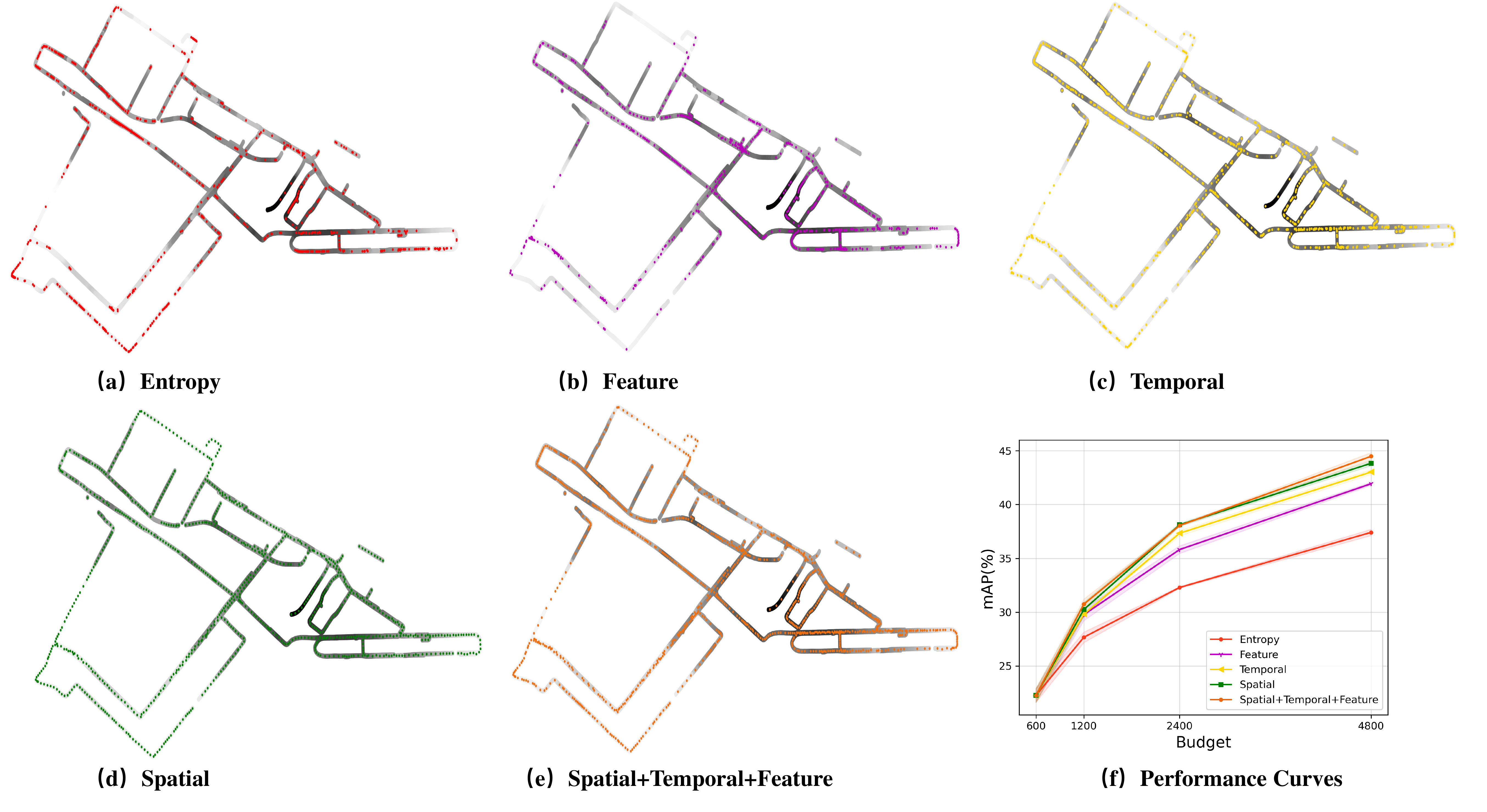}
        \caption{{Distribution of samples selected by different active learning strategies and the corresponding performance - (a) {Entropy}, (b) Feature Diversity, (c) Temporal Diversity, (d) Spatial Diversity and (e) Combined Spatial and Temporal Diversity. (f) Performance curves on nuScenes. Colors represent different selection strategies. The black-white color scale along the path represents original data distribution density (black indicates high density). The proposed spatial and temporal diversity is able to select a diverse of samples  without the need of a good feature extractor, while uncertainty-based sampling tends to concentrate in regions of high data density, resulting in redundant samples being selected.}}\label{FigHead}
\end{figure*}

{Active learning has demonstrated success across various computer vision tasks, including image classification \cite{gal2015bayesian, gal2017deep, li2017dropout}, 2D object detection \cite{aghdam2019active,roy2018deep,kao2018localization}, and semantic segmentation \cite{mackowiak2018cereals, cai2021exploring, cai2021revisiting}. However, little attention has been given to AL for 3D object detection. Feng et al. \cite{feng2019deep} made the first attempt by employing uncertainty-based sampling to query unlabeled samples, but their method is tightly coupled with frustum-based detectors. As a result, a general AL approach that can be applied universally to existing detectors is still missing. Additionally, diversity-based AL was not explored in \cite{feng2019deep}, leaving questions about whether and how multimodal information, typically available in AV datasets, can be leveraged to query informative samples.} %Our work addresses these gaps.}

% Active learning has proven successful across numerous computer vision tasks, including image classification \cite{gal2015bayesian, gal2017deep, li2017dropout}, 2D object detection \cite{aghdam2019active,roy2018deep,kao2018localization}, and semantic segmentation \cite{mackowiak2018cereals, cai2021exploring, cai2021revisiting}.}
% Little attention has been paid to AL for 3D object detection, and Feng et al. \cite{feng2019deep} made the first attempt by employing uncertainty-based sampling to query unlabeled samples. However, the method developed in \cite{feng2019deep} is entangled with frustum-based detectors, and a general AL method that can be universally applied to all existing detectors is lacking. Furthermore, diversity-based AL is not studied in \cite{feng2019deep}, and it is unclear whether and how the multimodal information that is typically available in an AV dataset can be utilized for querying informative samples. Our work fills this gap.

{In this work, we focus on diversity-based active learning (AL) for 3D object detection and propose that enforcing diversity in both spatial and temporal domains leads to superior performance. Our approach is motivated by the observation that data collected by autonomous vehicles at different locations and time steps correspond to distinct scenes and visual content. By selecting samples that vary in location and time, we can ensure a diverse set of samples that cover a wide range of object categories. Figure~\ref{FigHead} illustrates the behavior of different AL strategies and their corresponding performance curves. Moreover, we argue that methods traditionally considered effective for active learning in generic classification tasks—such as uncertainty sampling~(e.g. maximal entropy) and feature diversity—are not optimal for selecting informative samples in the autonomous driving scenario. These approaches tend to select redundant frames, and quantifying feature diversity in detection tasks is non-trivial. In contrast, the proposed spatial and temporal diversity effectively selects a diverse set of informative samples, and combining the spatial, temporal and feature diversities objectives results in the best detector performance within the same budget.}

{Another important consideration when applying active learning (AL) to 3D object detection is how to accurately measure annotation cost. Previous studies have measured this cost either by the number of annotated frames \cite{kao2018localization, aghdam2019active, yuan2021multiple} or by the number of annotated bounding boxes \cite{desai2020towards, gao2020consistency}. The former is imprecise, as different frames can contain varying numbers of objects, leading to significantly different annotation costs. The latter overlooks the fact that even nearly empty frames require time for the annotator to search for objects of interest. To address these limitations, we consider a more realistic scenario that accounts for both the cost of annotating a frame and the cost of annotating a 3D bounding box.}

% {Lastly, we tackle the cold-start problem in AL, where the first batch of samples is randomly selected due to the absence of a trained model for active selection \cite{gao2020consistency}. This often leads to suboptimal performance. We demonstrate that our proposed acquisition function can effectively provide a warm start to the AL cycles, mitigating this issue.}

% An important consideration when applying AL for 3D object detection is how to measure the annotation cost. Previous works have measured the annotation cost either by the number of annotated frames \cite{kao2018localization, aghdam2019active, yuan2021multiple} or the number of annotated bounding boxes \cite{desai2020towards, gao2020consistency}. The former is inaccurate, as different frames contain different numbers of objects and can have significantly different annotation costs. The latter ignores the fact that even an almost empty frame takes time for the annotator to look for objects of interest. Therefore, we consider a more realistic setting where both the cost of annotating a frame and a 3D bounding box are taken into account.

% Finally, we address the cold-start problem of AL, which refers to the first batch of samples being randomly selected (as a trained model is not available for active selection of the initial batch) \cite{gao2020consistency}. This often results in a sub-optimal solution. We show that the proposed acquisition function can effectively provide a warm-start for the AL cycles.

Our contributions can be summarized as below:

\begin{itemize}
\item We propose a novel diversity-based acquisition function for AL in 3D object detection. We exploit the multimodal information provided in an AV dataset and propose spatial and temporal diversity objectives for querying informative samples.
\item We propose to evaluate the AL methods under realistic annotation cost measurement. We show that the choice of cost measurement can significantly affect the evaluation results and highlight the importance of using realistic cost measurement to properly evaluate AL methods.%\cll{any insights we can provide for this?i.e., what kind of samples are most cost-effective?}\lzh{we did not any insight about it.}
% \item We look into the cold-start problem of AL and demonstrate that the proposed acquisition function can provide an effective solution to this.
\item We present the first pilot study of active learning for object detection in the context of autonomous driving tasks. We hope this work will draw attention from the community developing active learning algorithms to tackle realistic challenges.
\end{itemize}

\section{Related Work}
\subsection{Active Learning for Deep CNNs}
The success of deep CNNs on visual recognition tasks relies on the availability of large-scale annotated datasets. Active learning offers a promising solution to alleviate annotation efforts and there is a resurgence of interest in AL in the deep learning era \cite{ren2020survey}. Acquisition function is the core component in AL, and various AL strategies can be generally categorized into uncertainty-based and diversity-based methods. Uncertainty-based AL \cite{roth2006margin, joshi2009multi, houlsby2011bayesian, yang2015multi, yoo2019learning, fuchsgruber2024uncertainty} select the most uncertainty samples to label and different methods vary in how the uncertainty is estimated. \cite{beluch2018power} obtained the uncertainty through the estimation of an ensemble of classifiers. Learning-Loss\cite{yoo2019learning} proposed to use the pseudo loss as an indication of uncertainty.{~\cite{safaei2024entropic} proposed an Entropic Open-set AL (EOAL) framework. This framework employs entropy-based metrics to assess both known and unknown class distributions, allowing for the selection of informative samples during active learning rounds. ~\cite{li2024deep} introduced a novel algorithm designed to estimate data uncertainty by leveraging noise stability. The core concept of the algorithm involves measuring the deviation in model output from the original observation when the model parameters are subjected to random perturbations by noise. } Diversity-based AL \cite{sener2017active, hasan2015context, lin2017active, tang2019self,sinha2019variational} select the most diverse set of samples to label. Core-Set \cite{sener2017active} formulated the AL problem as minimizing the core-set loss and proposed solutions to minize the upper-bound of the core-set loss. VAAL \cite{sinha2019variational} trained an auto-encoder in an adversarial manner and select unlabeled samples that are most dissimilar with labeled ones based on the discriminator score. In the context of 3D point cloud deep learning, active learning was attempted by combining feature and spatial diversity~\cite{shi2021label,wu2021redal}. Our method is based on the Core-Set framework, and we propose novel diversity objectives to complement the traditionally used feature diversity.

\vspace{0.1cm}
\subsection{Active Learning for Object Detection}
Several attempts have been made to apply AL for object detection tasks, and most of them are focused on 2D object detection task. An image-level score was proposed by aggregating pixel-wise scores for active selection~\cite{aghdam2019active}. Instead of aggregating pixel-wise scores, \cite{brust2018active} investigate various methods to aggregate bounding box scores. {~\cite{blad2023} introduced a box-level active detection framework designed to control a box-based budget per cycle. Within this proposed box-level setting, a novel pipeline called the Complementary Pseudo Active Strategy (ComPAS) is developed. This strategy leverages both human annotations and model intelligence in a complementary manner. An efficient input-end committee queries labels only for informative objects, while simultaneously, well-learned targets are identified by the model and supplemented with pseudo-labels. ~\cite{park2023active} introduced a novel method called Hierarchical Uncertainty Aggregation (HUA), which aims to assess the informativeness of images by calculating the epistemic uncertainty associated with each bounding box.}  The disagreement of layer-wise prediction scores~\cite{roy2018deep} was utilized for query and the method is specific to one-stage object detectors with similar architecture of SSD \cite{liu2016ssd}. Following these, \cite{kao2018localization} combined the uncertainty in both classification and location prediction to query samples. {More recently, A two-stage approach was introduced by estimating the uncertainty at image level and selecting diverse samples~\cite{wu2022entropy}. \cite{yu2022consistency} introduced an active learning strategy based on the consistency between the original and augmented data. \cite{kothawade2022talisman} focused on selecting rare samples and proposed to efficiently target and acquire images with rare slices. A relevant problem of selecting most informative samples in the target domain was proposed to facilitate domain adaptation for object detection~\cite{yuan2023bi3d}.} Despite many attempts to address active learning for object detection tasks, these generic active learning methods for object detection are specifically optimized for 2D object detection and are entangled with the architecture of 2D object detectors, and thus a direct application to 3D object detection is either infeasible or sub-optimal. The only work we are aware that tackles AL for 3D object detection is reported in \cite{feng2019deep}, but the method is entangled with frustum-based detectors and only uncertainty-based AL is investigated. {More importantly, these methods do not consider the annotation costs associated with creating autonomous driving dataset where both frame-wise and object-wise costs are calculated towards the final cost. In this work, we consider a backbone agnostic active learning strategy and specifically optimize the method by considering the realistic annotation costs.}

\begin{comment}
Comparing with classification and semantic segmentation tasks, the first challenge that the detection task faces is to handle the inconsistency between the number of samples and the number of bounding boxes, which determines the measurement of annotation cost. To handle this challenge, existing methods directly use the image-based measurement\cite{brust2018active, aghdam2019active, kao2018localization} or box-based measurement\cite{desai2020towards, elezi2021towards} to measure the annotation cost.
Such inconsistency causes another problem is that existing a gap between the representation of bounding boxes and the representation of samples.
To bridge the gap, Brust \textit{et} al.\cite{brust2018active} considered different aggregation strategies and handled the selection selection imbalances to obtain the more comprehensive semantic uncertainties, Kao \textit{et} al.\cite{kao2018localization} proposed the localization-aware as the metrics for the localization uncertainty, Yuan \textit{et} al.\cite{yuan2021multiple} proposed the MI-AOD to bridge the gap between instance uncertainty and image uncertainty.
While promising, the above methods can not be migrated in 3D object detection effectively. For AL in 3D object detection, \cite{feng2019deep} leveraged 2D detector and images to assist AL. However, such method is only suitable for the frustum-based detectors. Comparing with it, our proposed method is applicable to all existing 3D detectors.
\end{comment}

\vspace{0.1cm}
\subsection{3D Object Detection}
Existing 3D object detectors can be categorized as either point-based or voxel-based methods, depending on how the point cloud data is represented. Point-based detectors \cite{shi20193d, yang20203dssd, wang2019frustum, qi2018frustum, qi2019deep} use a point-wise feature extractor \cite{qi2017pointnet, qi2017pointnet++, wang2019dynamic} as a backbone and directly extract point-wise features from the input point cloud. Proposals are generated based on the position of the point cloud, which can provide more fine-grained position information. However, point-based methods typically downsample the input point cloud to a fixed number of points, which may result in the loss of some input information.
On the other hand, voxel-based detectors \cite{zhou2018voxelnet, yan2018second, zhu2019class, yin2021center, chen2020object, chen2020every} quantize the input point cloud as regular voxels and utilize 3D CNNs to operate directly on the voxels. The extracted 3D feature map is then projected into a bird's-eye-view (BEV) feature map, and proposals are generated from the projected 2D feature map. { Among these methods, UniMODE~\cite{li2024unimode} proposed a novel detector grounded in the BEV detection paradigm. This approach leverages explicit feature projection to effectively address the challenges of geometry learning ambiguity that arise when training detectors across multiple scenarios. UniMODE marks the first successful generalization of a BEV detector for unified 3D object detection. ~\cite{li2024bevnext} introduced BEVNeXt, a fully enhanced dense BEV framework for multi-view 3D object detection. BEVNeXt addresses the limitations of existing dense BEV-based 3D object detectors by introducing several novel components. These include a CRF-modulated depth estimation module that enforces object-level consistency, a long-term temporal aggregation module with extended receptive fields, and a two-stage object decoder that combines perspective techniques with CRF-modulated depth embedding.} While voxelization can result in the loss of fine-grained position information, it has the advantage of establishing regularities for the irregular point cloud, and allows for the use of powerful 3D CNNs to extract features. 
In our experiments, we employ VoxelNet \cite{zhou2018voxelnet}, as the detection model. {We also evaluate on BEVFusion~\cite{liu2023bevfusion} which exploits both RGB and point cloud modalities to investigate the generalization ability of the proposed method.}

% \noindent\textbf{Classical Active Learning}
% Uncertainty and diversity are the two most widely used objectives to optimize for active learning(AL).
% Uncertainty-based methods\cite{lewis1994heterogeneous, joshi2009multi, roth2006margin, settles2007multiple, luo2013latent} characterize the informativeness by high model uncertainty on the sample, and aim to establish a reasonable criterion to measure the uncertainty. They select the top-K informative samples from the unlabeled pool.
% Diversity-based methods\cite{sener2017active, hasan2015context, lin2017active, tang2019self} formulate the AL problem as a diversification problem and evaluate the diversity using distances between selected samples. It is crucial to establish a reliable distance criterion. Different from the uncertainty-based methods, diversity-based methods consider the relation and difference between unlabeled pool and labeled pool, which leads to avoid overlap between similar samples.

% Existing works do not focus on the 3D object detection in autonomous driving and do not exploit the properties in AV dataset.

\vspace{0.1cm}

% \begin{figure}[!tb]
%     \centering
%     \includegraphics[width=1.0\linewidth]{images/spatial_temporal.pdf}
%     \caption{Illustration of the spatial and temporal distributions of samples(at an intersection in boston-seaport) - (a) spatial distribution, (b) temporal distribution. Points with different colors represent different samples, and the points with edge of the red dashed line represent the samples selected by our proposed selection strategy. $\rightarrow$ represents the same interval on the timeline.}
%     \label{FigSpatialTemporal}
% \end{figure}

\section{Methodology}
\begin{figure*}[tb]
    \centering
    \includegraphics[width=0.99\linewidth]{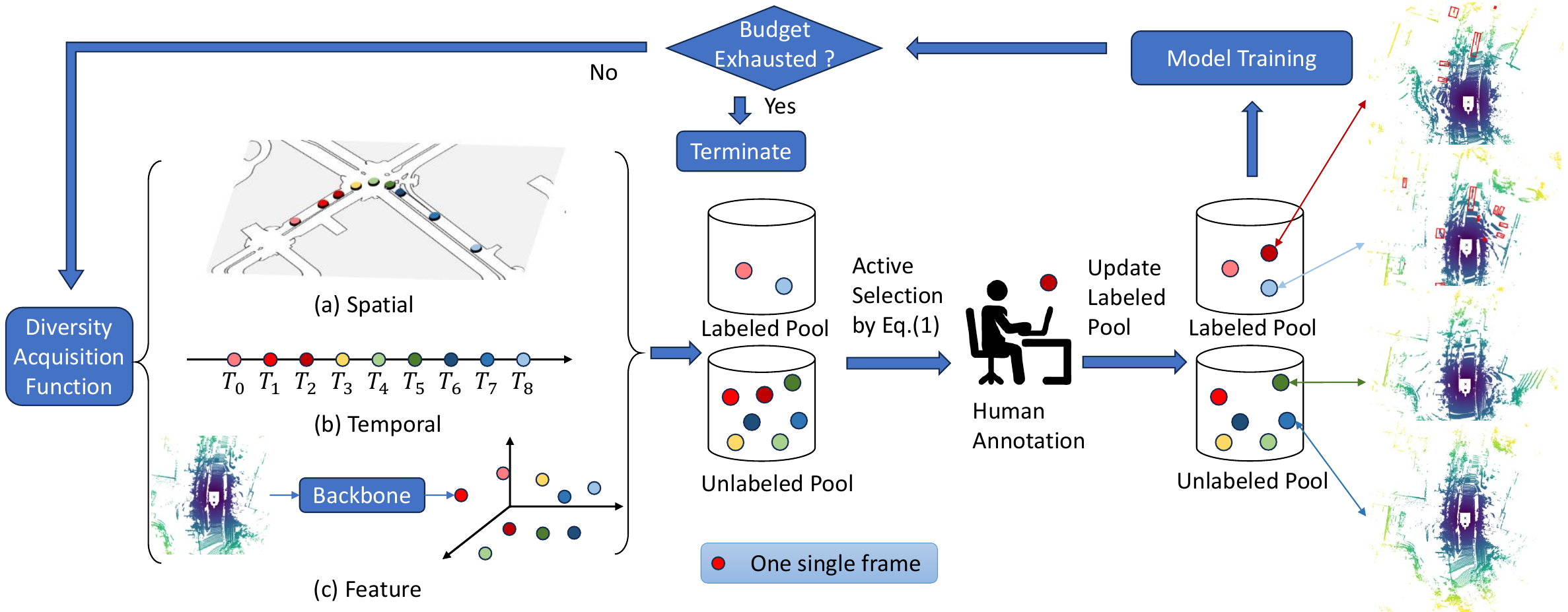}
    %\caption{The proposed AL pipeline for 3D object detection. The spatial, temporal and feature diversity terms are complementary to each other.}
    %\caption{The proposed AL pipeline for 3D object detection. (a) spatial distribution(at an intersection in boston-seaport), (b) temporal distribution, (c) feature distribution. Points with different colors represent different samples. $\rightarrow$ in (b) represents the same interval on the timeline. The vehicle velocity causes the inconsistency between spatial distribution and temporal distribution. The spatial diversity and temporal diversity are complementary.}
    \caption{{We illustrate the pipeline of active learning for 3D object detection on 9 sample frames $T_0$ to $T_8$. For all available data, (a) Spatial and (b) Temporal information are directly extracted from metadata. (c) Feature information is computed as the global average pooling of the feature map  produced by the backbone network. We first initialize the active learning cycle with limited labeled data and large unlabeled data. Unlabeled samples are then chosen by Eq.~\ref{EqKCenter} for annotation. The annotated sample is appended to the labeled pool for model training. The cycle is repeated until labeling budget is exhausted.}}
    \label{FigPipeline}
\end{figure*}

The overview of the proposed AL pipeline for 3D object detection is presented in Fig.~\ref{FigPipeline}. AL is an iterative process in which an informative subset of samples is selected based on an acquisition function in each cycle. The selected samples are then sent to an oracle for annotation, and the model is updated on all the annotated samples so far. The process is repeated until the annotation budget is exhausted. First, we introduce the diversity-based AL framework. We then provide a detailed description of our proposed acquisition function that incorporates novel spatial and temporal diversity objectives. Finally, we present a realistic method for measuring annotation cost.

\subsection{Diversity-based AL}\label{SubSecOverview}
\begin{figure}
    \centering
    \includegraphics[width=0.99\linewidth]{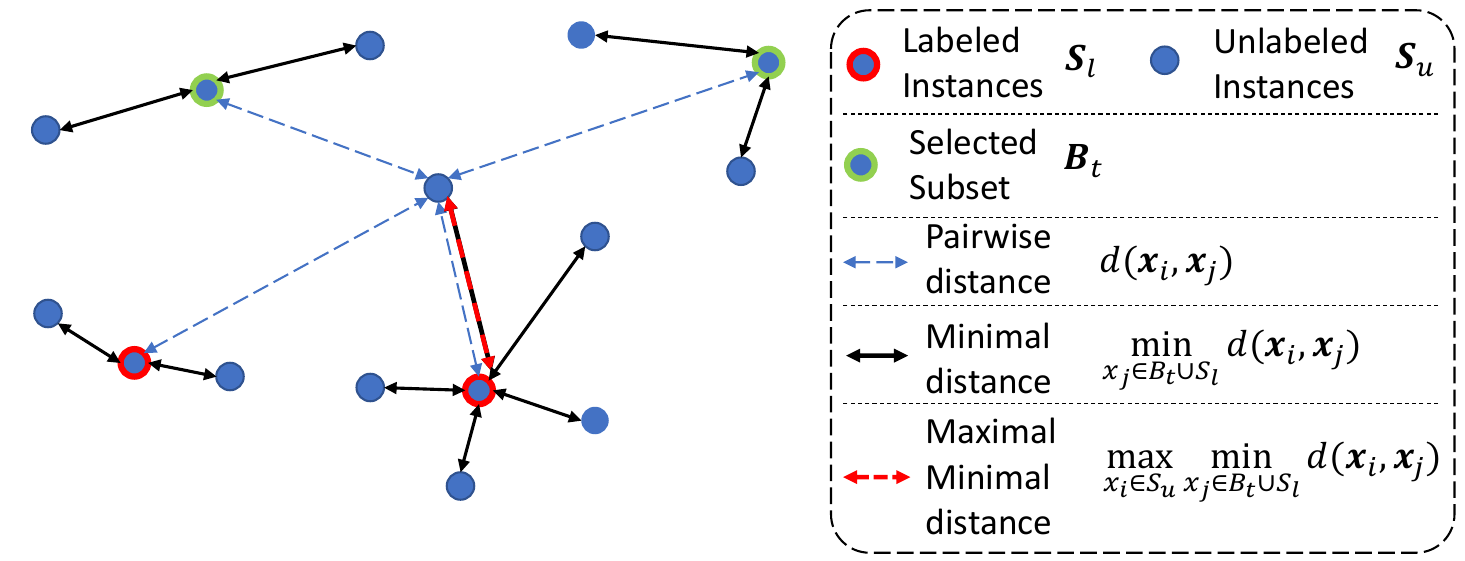}
    \caption{\textcolor{black}{An illustration of the k-Center objective. Two instances (red outlined circles) are labeled and another two instances (green outlined circles) are the candidates to be selected. The distance indicated by red line is the maximal minimal distance to be optimized by Eq.~\ref{EqKCenter}. Choosing which two candidate instances (green outlined circles) are the optimization variables.}}
    \label{fig:kcenter}
\end{figure}

Diversity-based active learning aims to select a diverse subset of samples that can represent the distribution of the unlabeled pool. One of the state-of-the-art methods for diversity-based AL is Core-Set, proposed by Sener and Savarese \cite{sener2017active}. Core-Set formulates the AL problem as minimizing the core-set loss, which involves selecting a set of points such that the difference between the average empirical training loss on this subset and the average empirical loss on the entire dataset is minimized. The authors showed that minimizing the upper bound of the core-set loss is equivalent to solving a k-Center problem. Mathematically, denoting the unlabeled pool as $\set{S}_u=\{\matr{x}_i\}_{i=1\cdots N_u}$ and the labeled pool as $\set{S}_l=\{\matr{x}_j\}_{j=1\cdots N_l}$, at each AL cycle $t$, we aim to select a subset $\set{B}_t$ from $\set{S}_u$ by solving the k-Center objective:

\begin{equation}\label{EqKCenter}
    \min_{\set{B}_t\subseteq\set{S}_u} \max_{\matr{x}_i \in\set{S}_u} \min_{\matr{x}_j \in \set{B}_t\cup\set{S}_l} d(\matr{x}_i, \matr{x}_j), \quad s.t.\quad \mathcal{C}(\set{B}_t)\leq b_t,
\end{equation}

where $d(\cdot, \cdot)$ is a distance measure between two samples, $\mathcal{C}(\cdot)$ is the annotation cost for labeling a subset, and $b_t$ is the annotation budget for cycle $t$. 
\Rev{The general idea of the K-Center objective in Eq.~\ref{EqKCenter} is to select a subset of frames $\set{B}_t$ from the unlabeled pool $\set{S}_u$ such that the maximal distance of any unlabeled instance to the closest selected instance is minimized. A graphical illustration is provided in Fig.~\ref{fig:kcenter}. Solving Eq.~\ref{EqKCenter} is NP-hard and we choose a greedy algorithm, achieving an approximation factor of 2, by iteratively selecting the farthest sample into the subset $\set{B}_t$ until the budget is exhausted~\cite{har2011geometric}. The algorithm is summarized in Algorithm~\ref{AlgKCenterGreedy}.}

% Solving Eq.~(\ref{EqKCenter}) is NP-hard, and we opt for a greedy solution for AL selection, which is summarized in Algorithm~\ref{AlgKCenterGreedy}.

\renewcommand{\algorithmicrequire}{ \textbf{Input: }} %Use Input in the format of Algorithm
\renewcommand{\algorithmicensure}{ \textbf{Output:}} %UseOutput in the format of Algorithm
\begin{algorithm}[h]
  \caption{Greedy solution for AL selection}
  \label{AlgKCenterGreedy}
  \begin{algorithmic}[1]
        \Require Budgets $\{b_0,\ b_1,\ \ldots,\ b_{T-1}\}$; Dataset $\set{S}$
        \State Initialize the label set $\set{S}_l$ with $\mathcal{C}(\set{S}_l)\leq b_0$
        \State Initialize the unlabeled set $\set{S}_u=\set{S} \setminus \set{S}_l$

        \For {$t=1,\ \ldots,T-1$}
            \State $\set{B}_t=\emptyset$\
            \State $\tilde{b}$ = 0
            \While{$\tilde{b} < b_t$}
                \State $\tilde{\matr{x}} = \arg \max_{\matr{x}_i \in \set{S}_u} \min_{\matr{x}_j \in \set{S}_l \cup \set{B}_t } d(\matr{x}_i, \matr{x}_j)$
                \State $\set{B}_t \leftarrow \set{B}_t \cup \tilde{\matr{x}}$
                \State $\set{S}_u \leftarrow \set{S}_u \setminus \tilde{\matr{x}}$
                \State $\tilde{b} \leftarrow \mathcal{C}(\set{B}_t)$
            \EndWhile
            \State Annotate the samples in $\set{B}_t$
            \State $\set{S}_l \leftarrow \set{S}_l \cup \set{B}_t$
            \State Train detector $\Phi_t$ on $\set{S}_l$
        \EndFor
        \State \Return Detector $\Phi_{T-1}$;
    \end{algorithmic}
\end{algorithm}

A common practice to measure $d(\matr{x}_i, \matr{x}_j)$ is to embed the samples in a feature space using a feature extractor $\Theta$ and compute the distance by $L_p$ norm:

\begin{equation} \label{EqDf}
    d(\matr{x}_i, \matr{x}_j)=||\Theta(\matr{x}_i)-\Theta(\matr{x}_j)||_p.
\end{equation}
However, it is non-trivial to design a good feature extractor for the task. In the following section, we explore how the multimodal information in an AV dataset can be utilized to design more effective diversity terms.

\subsection{Enforcing Diversity in Spatial, Temporal and Feature Space}\label{SubSecEnforce}
Existing AV datasets comprise multimodal data including images, point clouds, and GPS/IMU data. The multimodal data is complementary to each other, and how to combine multimodal measurements in a principled manner for robust detection has attracted much attention from the AV community. While most works focus on fusing image and point cloud data for 3D object detection \cite{chen2017multi, Vora_2020_CVPR}, we are interested in exploiting the GPS/IMU data for active learning of 3D object detectors.

The GPS/IMU system provides accurate location information for the vehicle. Intuitively, different locations traversed by the autonomous vehicle correspond to different scenes and visual content, and by enforcing diversity in the location of the selected samples, we can obtain a diverse set of samples covering different object categories without relying on a good feature extractor. This motivates us to design a spatial diversity objective for the k-Center optimization. Furthermore, we observe that an autonomous driving vehicle may visit the same place at different times for data collection, and the data collected at the same location but different times vary in traffic and weather conditions. We thus propose to enforce diversity in temporal space to complement the spatial diversity for optimal sample selection. Finally, image-wise feature characterizes the visual patterns in a scene and selecting visually diverse scenes to annotate provides additional guarantees on selecting diverse samples. In the following of this section, we provide more detailed definitions of the proposed spatial, temporal and feature diversity objectives.

\noindent\textbf{Spatial Diversity Objective}:
As the multi-modal sensors are synchronized during data collection, each sample in $\set{S}_u$ is associated with a spatial location. Denoting the location of sample $\matr{x}_i$ as $loc(\matr{x}_i)$, we measure the spatial distance between two samples by manifold distance. Specifically, we first build a KNN graph with a distance matrix $\matr{D}\in \mathbb{R}^{(N_l+N_u) \times (N_l+N_u)}$ defined as follows:%\cll{why manifold distance is better than Euclidean distance? some motivation goes here. also, ablation study should provided in Experiments}\lzh{I did the experiments about the comparison betweean Euclidean and manifold, unfortunately, their performance are not much different . I discussed this question with Alex, and we agree that just saying we use the manifold distance, because the manifold is more intuitive and realistic. We can not provide the experiments to prove the manifold is much better than Euclidean}
\begin{equation}
    d_{ij}=
    \begin{cases}
    ||loc(\matr{x}_i) -loc(\matr{x}_j)||_2^2, & \text{if } \matr{x}_i \in NN_k (\matr{x}_j ) \\
    +\infty, & \text{otherwise}
    \end{cases}
\end{equation}
% TODO in coding
where $NN_k(\cdot)$ is the set of k-nearest neighbors for a sample in the Euclidean space. The spatial distance between $\matr{x}_i$ and $\matr{x}_j$ is then defined as their shortest path on the KNN graph. This problem can be efficiently solved by Dijkstra's algorithm:
\begin{equation}
\begin{split}\label{EqDs}
    &d_s(\matr{x}_i, \matr{x}_j) =\min_{\{c_{pq}\}}\sum_{p,q=1, \cdots, (N_l+N_u)}d_{pq}c_{pq}\\
    &s.t.\quad c_{pq}\geq0\text{ and } \forall p,\sum_q c_{pq}-\sum_q c_{qp}=
    \begin{cases}
    1, \; \text{if } p=i\\
    -1,\; \text{if } p=j\\
    0,\; \text{otherwise}
    \end{cases}
\end{split}
\end{equation}
For samples collected from different areas, e.g. one sample collected from Singapore vs. another collected from Boston, we define the spatial distance as a large constant.% \cll{what is the value of this constant?}\lzh{Infinite or $1e^6$, they lead to the same sampling results.}

\noindent\textbf{Temporal Diversity Objective}:
A sample in an AV dataset is also associated with a data stream id $si\in \mathbbm{N}$ and the timestamp $ts\in \mathbbm{R}$. We define the temporal distance as below,
\begin{equation}\label{EqDt}
    d_t(\matr{x}_i, \matr{x}_j) = 
    \begin{cases}
        |ts(\matr{x}_i) - ts(\matr{x}_j)|,\ \text{if}\ si(\matr{x}_i) = si(\matr{x}_j) \\
        +\infty. \ \text{otherwise}
    \end{cases}
\end{equation}

\noindent\textbf{\textcolor{black}{
Feature Diversity Objective}}:
\textcolor{black}{
Specifically, for the given data sample $\matr{x}_i$, we use the backbone of predictor $\Theta$ to extract the feature map $\matr{F}_i$ from the point cloud of $\matr{x}_i$, and obtain the representation feature vector $\bm{f}_i$ by aggregating feature map $\matr{F}_i$ through global average pooling. After obtaining the representing feature vector $\bm{f}_i$, we can embed the data sample into the feature space.
Given two arbitrary sample $\matr{x}_i$ and $\matr{x}_j$, the distance $d_f$ can be defined as the Euclidean distance between representation features.
\begin{equation}\label{eq:feature}
    d_f(\matr{x}_i,\matr{x}_j) = ||\bm{f}_i-\bm{f}_j||^2_2
\end{equation}
}

\subsection{Combining Diversity Terms}\label{SubSecAcquisition}
The various diversity terms defined above, such as spatial, temporal, and feature diversity, are complementary to each other, and we can combine them to achieve better performance. Since the values for each diversity objective are on different scales, we first normalize them to the [0, 1] range using a radial basis function (RBF) kernel. Specifically, for the spatial diversity term $d_s(\matr{x}_i, \matr{x}_j)$, we normalize it using the following equation:
\begin{equation}\label{EqRbf}
\bar{d}_{s}(\matr{x}_i, \matr{x}_j) = 1 - \exp (-d_s(\matr{x}_i, \matr{x}_j)).
\end{equation}
The temporal diversity term $d_t$ and $d_f$ can be normalized in a similar way. The distance metric used in Algorithm~\ref{AlgKCenterGreedy} is then defined as:
\begin{equation}\label{EqDagree}
    d(\matr{x}_i, \matr{x}_j) = \lambda_s \bar{d}_{s}(\matr{x}_i, \matr{x}_j) + \lambda_t \bar{d}_{t}(\matr{x}_i, \matr{x}_j) + \lambda_f \bar{d}_{f}(\matr{x}_i, \matr{x}_j).
\end{equation}

\subsection{Annotation Cost Measurement}\label{SubSecMeasurement}
In contrast to previous works that use either the number of annotated frames or the number of annotated bounding boxes to measure annotation cost, we propose to take into consideration the cost of annotating both. Specifically, the annotation cost $\mathcal{C}(\set{B}_t)$ used in Eq.~(\ref{EqKCenter}) and Algorithm~\ref{AlgKCenterGreedy} is computed as:
\begin{equation}\label{EquCost}\
    \mathcal{C}(\set{B}_t)=c_f * n_f({\set{B}_t}) + c_b * n_b({\set{B}_t}),
\end{equation}
where $c_f$ and $c_b$ is the cost for annotating one frame and one bounding box, respectively, and $n_f$ and $n_b$ denote the total number of frames and bounding boxes in the selected samples. The actual value for $c_f$ and $c_b$ varies among different vendors. In the experiments, we determine the values based on discussions with the nuScenes team and investigate the influence of different settings in Section~\ref{sec:influence_cost}.

\section{Experiment}
\subsection{Experimental Settings}
\noindent{\textbf{Evaluation Datasets}}:
We perform experiments on the nuScenes\cite{Caesar_2020_CVPR}, which is a popular multimodal dataset used for benchmarking 3D object detection algorithms. The training set of nuScenes contains 28,130 LiDAR point clouds and 1,255,109 bounding boxes, and the validation set of nuScenes contains 6,019 LiDAR point clouds. The nuScenes contains 10 categories, namely: car, truck, construction vehicle, bus, trailer, barrier, motorcycle, bicycle, pedestrian, traffic cone and ignore. We perform active selection on the training set and report the detector performance on the validation set. 

\noindent{\textbf{Evaluation Metrics}}:
Following the official nuScenes evaluation protocol, we report mean Average Precision(mAP)~\cite{Caesar_2020_CVPR} as the evaluation metric to compare different methods. {All experiments are reported the mean the standard deviation across 3 runs with different random seeds.}

\noindent{\textbf{Detection Model}}:
We employ the open-source VoxelNet\cite{zhou2018voxelnet} as our 3D object detection detector in all experiments.
We voxelize the input point clouds in the range of $[-51.2, 51.2]$, $[-51.2, 51.2]$ and $[-5.0, 3.0]$ for axis x, y, z, respectively. {Additionally, we evaluate a stronger detection backbone, BEVFusion~\cite{liu2023bevfusion}, which fuses RGB and point cloud modalities for improved detection results, for comparing the overall active learning performance.}

\noindent{\textbf{Fully Supervised Baseline}}:
We train a detector using the entire training set, which serves as a performance upper bound for all AL methods. We follow the default training setting in VoxelNet\cite{zhou2018voxelnet}. For nuScenes, we trained for a total 20 epoches with a batch size of 8. The mAP of the fully supervised baseline is 51.00\%.

\noindent{\textbf{Implementation Details}}:
We perform sample selection at the granularity of frame, and once a frame is selected, all the objects of interest within the frame are annotated. We set $c_f=0.12,\ c_b=0.04$ in Eq.~ \ref{EquCost}. With this setting, the budget for annotating the entire training set in nuScenes is $36,314$. We conduct AL at the budget of 600/1200/2400/4800, where the first 600 labeled samples are randomly selected and fixed for all competing methods. The number of nearest neighbor used in constructing the KNN graph for computing spatial diversity is set to $K=8$.
The training hyper-parameters remain the same as the fully supervised baseline. {After labeling a each batch of data, we train the detection model from scratch following the standard setup reported in \cite{zhou2018voxelnet} with 20 epochs on LiDAR data for VoxelNet~\cite{zhou2018voxelnet} and additional 6 epochs on RGB data for BEVFusion~\cite{liu2023bevfusion}. We use a server with 4 RTX3090 GPUs for training. It roughly takes 2.5 hours and 2 hours for training with every 1200 labeled samples with VoxelNet and BEVFusion respectively on this platform. }
{For all experiments, we conduct three independent runs with different random seeds and report the averaged results.}

\subsection{Experimental Results}
\label{sec:exp_results}
\noindent
\textbf{Comparing the State-of-the-Arts Methods}:
We compared with three state-of-the-art active learning methods by adapting both generic active learning methods and active learning method for object detection to autonomous driving detection task. Specifically, the following methods are benchmarked:

\noindent
\textbf{Badge~\cite{ash2019deep}:} This strategy considers both model uncertainty and feature for data acquisition. The gradient features are computed by accumulating the gradients of loss w.r.t. penultimate layers. We use the classification branch to calculate gradient features in our task. Samples are then selected by K-means++.

\noindent\textbf{CALD~\cite{yu2022consistency}:} This approach employs a simple idea that predictions upon different data augmentations by a good model should be consistent. Hence, a frame with predictions that do not agree between augmentations, measured by the J-S divergence between classification predictions and IoU between boxes, will be more likely to be selected.

\noindent\textbf{PPAL~\cite{yang2024ppal}:} This method is specifically tailored for object detection by considering both model uncertainty and feature diversity. Classification and localization difficulties are fused to re-weight model uncertainties. K-means++ is employed to select the final queries.

\noindent{\textbf{UWE~\cite{uwe2024}:} This approach was inspired by the effectiveness of combining model uncertainty and feature diversity for active learning. To enable the concept for more diverse computer vision tasks, UWE introduced a uncertainty re-weighted representation for active learning.}

\noindent\textbf{Comparing Different Diversity Terms}:
We consider three diversity terms, i.e. feature, spatial and temporal, to measure the distance between two samples. We set the random selection as baseline and investigate the effect of each diversity term and their combination on AL performance. In specific, we consider the following variants:

\noindent\textbf{Random:} This strategy randomly selects samples to annotate.

\noindent\textbf{Feature~\cite{sener2017active}:} This strategy uses Eq.~(\ref{eq:feature}) to measure sample distance. We obtain the feature of each sample by averaging the output feature map of the 3D feature extractor. %\cll{some explanation on how to extract features}

\noindent\textbf{Spatial~\cite{cai2021exploring}:} This strategy uses Eq.~(\ref{EqDs}) to measure sample distance.

\noindent\textbf{Temporal:} This strategy uses Eq.~(\ref{EqDt}) to measure sample distance.

\noindent\textbf{Spatial+Temporal~(Ours):} This strategy uses Eq.~(\ref{EqDagree}) with $\lambda_s=1, \lambda_t=1, \lambda_f=0$ to measure sample distance.

\noindent{\textbf{Spatial+Temporal+Feature~(Ours):}} This strategy uses Eq.~(\ref{EqDagree}) with $\lambda_s=1, \lambda_t=1, \lambda_f=1$ to measure sample distance.

\begin{table*}[]
\centering
    \setlength\tabcolsep{2pt} % default value: 6pt
     \caption{{Comparing state-of-the-art active learning methods and different diversity terms for active learning with \textbf{VoxelNet} and \textbf{BEVFusion} backbones. \textbf{Spatial}, \textbf{Temporal} and \textbf{Features} are shortened as \textbf{Spa.}, \textbf{Temp.} and \textbf{Feat.} respectively. {We report the average $\pm$ standard deviation of mAP from 3 random runs for each method. \textcolor{red}{Red}, \textcolor{blue}{blue} and \textcolor{green}{green} colors indicate the best, second best and the third best results respectively. All numbers except average ranking are in (\%).}}}\label{TableCompDiveristy}
    \resizebox{0.99\linewidth}{!}{%
\begin{tabular}{llllllllll}
\toprule
Backbone         & \multicolumn{4}{c|}{VoxelNet~\cite{zhou2018voxelnet}}                                                                                    & \multicolumn{4}{c}{BEVFusion~\cite{liu2023bevfusion}}  &       \\ \cmidrule{1-9} 
                 & \multicolumn{8}{c}{Budget}                                          \\ \cmidrule{2-9}
A. L. Methods    & \multicolumn{1}{c}{600} & \multicolumn{1}{c}{1200} & \multicolumn{1}{c}{2400} & \multicolumn{1}{c|}{4800}        & \multicolumn{1}{c}{600} & \multicolumn{1}{c}{1200} & \multicolumn{1}{c}{2400} & \multicolumn{1}{c}{4800} & Avg. Rank. \\ \hline
Random           & 22.27$\pm$0.63             & 28.76$\pm$0.01              & 36.16$\pm$0.43              & \multicolumn{1}{l|}{42.07$\pm$0.04} & 31.29$\pm$1.56             & 41.80$\pm$0.42              & 50.13$\pm$0.60              & 56.76$\pm$1.32      & \multicolumn{1}{c}{7.5}       \\
Badge~\cite{ash2019deep}            & 22.27$\pm$0.63             & 29.09$\pm$0.29               & 35.92$\pm$0.10               & \multicolumn{1}{l|}{42.93$\pm$0.04}  & 31.29$\pm$1.56             & \textcolor{green}{43.53$\pm$0.66}               & 51.61$\pm$1.39               & 57.43$\pm$1.56    & \multicolumn{1}{c}{5.2}           \\
CALD~\cite{yu2022consistency}             & 22.27$\pm$0.63             & 28.17$\pm$0.39               & 33.98$\pm$0.25               & \multicolumn{1}{l|}{39.42$\pm$0.28}  & 31.29$\pm$1.56             & 39.65$\pm$0.92               & 48.29$\pm$0.52               & 54.44$\pm$1.12       & \multicolumn{1}{c}{9.3}        \\
PPAL~\cite{yang2024ppal}             & 22.27$\pm$0.63             & 26.44$\pm$0.46               & 32.67$\pm$0.63               & \multicolumn{1}{l|}{39.65$\pm$0.20}  & 31.29$\pm$1.56             & 36.47$\pm$1.46               & 44.27$\pm$0.57               & 53.06$\pm$0.59     & \multicolumn{1}{c}{10.3}         \\
{UWE}~\cite{uwe2024} & 22.27$\pm$0.63 & 28.84$\pm$0.06 & 36.50$\pm$0.28 & \multicolumn{1}{l|}{43.20$\pm$0.64} & 31.29$\pm$1.56 & 42.58$\pm$1.45& 51.11$\pm$1.62 & 57.45$\pm$0.45& \multicolumn{1}{c}{5.3}\\
Entropy~\cite{holub2008entropy}          & 22.27$\pm$0.63             & 27.67$\pm$0.49              & 32.30$\pm$0.07              & \multicolumn{1}{l|}{37.36$\pm$0.29} & 31.29$\pm$1.56             & 40.65$\pm$1.54              & 47.27$\pm$1.01              & 52.03$\pm$0.92     & \multicolumn{1}{c}{10.3}        \\
Feature~\cite{sener2017active}          & 22.27$\pm$0.63             & 29.77$\pm$0.55              & 35.82$\pm$0.35              & \multicolumn{1}{l|}{41.94$\pm$0.13} & 31.29$\pm$1.56             & 43.12$\pm$0.57              & 50.71$\pm$0.42              & 56.98$\pm$1.66     & \multicolumn{1}{c}{6.3}         \\
Spatial~\cite{cai2021exploring}          & 22.27$\pm$0.63             & \textcolor{green}{30.25$\pm$0.68}              & \textcolor{red}{38.12$\pm$0.13}              & \multicolumn{1}{l|}{\textcolor{green}{43.84$\pm$0.22}} & 31.29$\pm$1.56             & 42.74$\pm$1.74              & \textcolor{red}{52.66$\pm$0.32}              & \textcolor{green}{57.92$\pm$1.13}     & \multicolumn{1}{c}{\textcolor{green}{2.7}}         \\
Temporal~(Ours)         & 22.27$\pm$0.63             & 29.81$\pm$0.05              & 37.34$\pm$0.23              & \multicolumn{1}{l|}{43.03$\pm$0.22} & 31.29$\pm$1.56             & 42.42$\pm$0.60              & 51.27$\pm$0.72              & 56.92$\pm$1.30     & \multicolumn{1}{c}{5.3}         \\
Spa.+Temp.~(Ours)       & 22.27$\pm$0.63             & \textcolor{red}{30.90$\pm$0.60}              & \textcolor{blue}{38.10$\pm$0.95}              & \multicolumn{1}{l|}{\textcolor{blue}{44.20$\pm$0.38}} & 31.29$\pm$1.56             & \textcolor{red}{44.06$\pm$1.28}              & \textcolor{blue}{52.66$\pm$1.59}              & \textcolor{blue}{58.12$\pm$1.16}     & \multicolumn{1}{c}{\textcolor{red}{1.7}}         \\
Spa.+Temp.+Feat.~(Ours) & 22.27$\pm$0.63             & \textcolor{blue}{30.75$\pm$0.48}              & \textcolor{green}{38.06$\pm$0.07}              & \multicolumn{1}{l|}{\textcolor{red}{44.49$\pm$0.35}} & 31.29$\pm$1.56             & \textcolor{blue}{43.71$\pm$0.95}              & \textcolor{green}{52.43$\pm$0.75}              & \textcolor{red}{58.24$\pm$1.08}     & \multicolumn{1}{c}{\textcolor{blue}{2.0}}         \\ \bottomrule
\end{tabular}
 }
\end{table*}

\noindent\textbf{\Rev{Analysis of Results}}:
\Rev{%For fair comparison, the first batch (budget=600) is randomly selected and kept the same for all methods. 
We present the results in Table~\ref{TableCompDiveristy}. {We report two metrics for each method. The mAP indicates the accuracy of object detection and the average ranking (Avg. Rank.) is the average ranking (lower the better) of each method across three budgets (1200, 2400, 4800) for both VoxelNet and BEVFusion backbones.} We make the following observations from the results. First, we notice that incorporating spatial and temporal diversity plays the most important role in selecting most informative samples. In particular, spatial diversity~(Spatial) alone performs comparatively to methods combining multiple diversity terms~(Spa.+Temp.) at budget 1200 and 2400, suggesting the importance of selecting diverse scenes. {When budget is increased to 4800, combining spatial, temporal and feature diversity demonstrates clear advantage over individual diversity terms, ranked in top 2 among all methods, suggesting the importance of considering multiple types of diversity for active selection.} We further notice that uncertainty-based active learning method~(Entropy) performs exceptionally worse than diversity-based methods. One drawback of uncertainty-based sampling comes from the observation that neural networks tends to predict similar output for similar input, and thus similar samples will have similar uncertainty values. Directly selecting the top-k uncertain samples will result in labeling redundant scenes. {The hypothesis is further strengthened by the relatively poor results of CALD, which considers model uncertainty alone, {and UWE, which integrates uncertainty into feature diversity}.} This problem is more severe on AV datasets, where the velocity of the autonomous vehicle is affected by the traffic condition, resulting in different sampling density at different parts of the path, e.g. slower velocity results in higher sampling density as the recording rate of the sensors is fixed. Regions with high sampling density usually correspond to busy scenes that contain a relatively large number of objects. The object detector has relatively low confidence (high uncertainty) on these scenes which will be picked up by uncertainty-based sampling. Hence, the high sampling density of these scenes will encourage uncertainty-based sampling to select more uncertain yet redundant samples, which may eventually harm the performance of object detector.} 
{Finally, among the three state-of-the-art methods, Badge and UWE achieve the top two results (average ranking 5.2 \& 5.3). We speculate that the advantage of Badge and UWE owes to considering both uncertainty and feature diversity. In contrast, CALD only considers sample uncertainty alone. In conclusion, combining both spatial and temporal diversity is the recommended way for active learning on AV dataset.}

\begin{figure}[tb]
    \centering
    \includegraphics[width=1.0\linewidth]{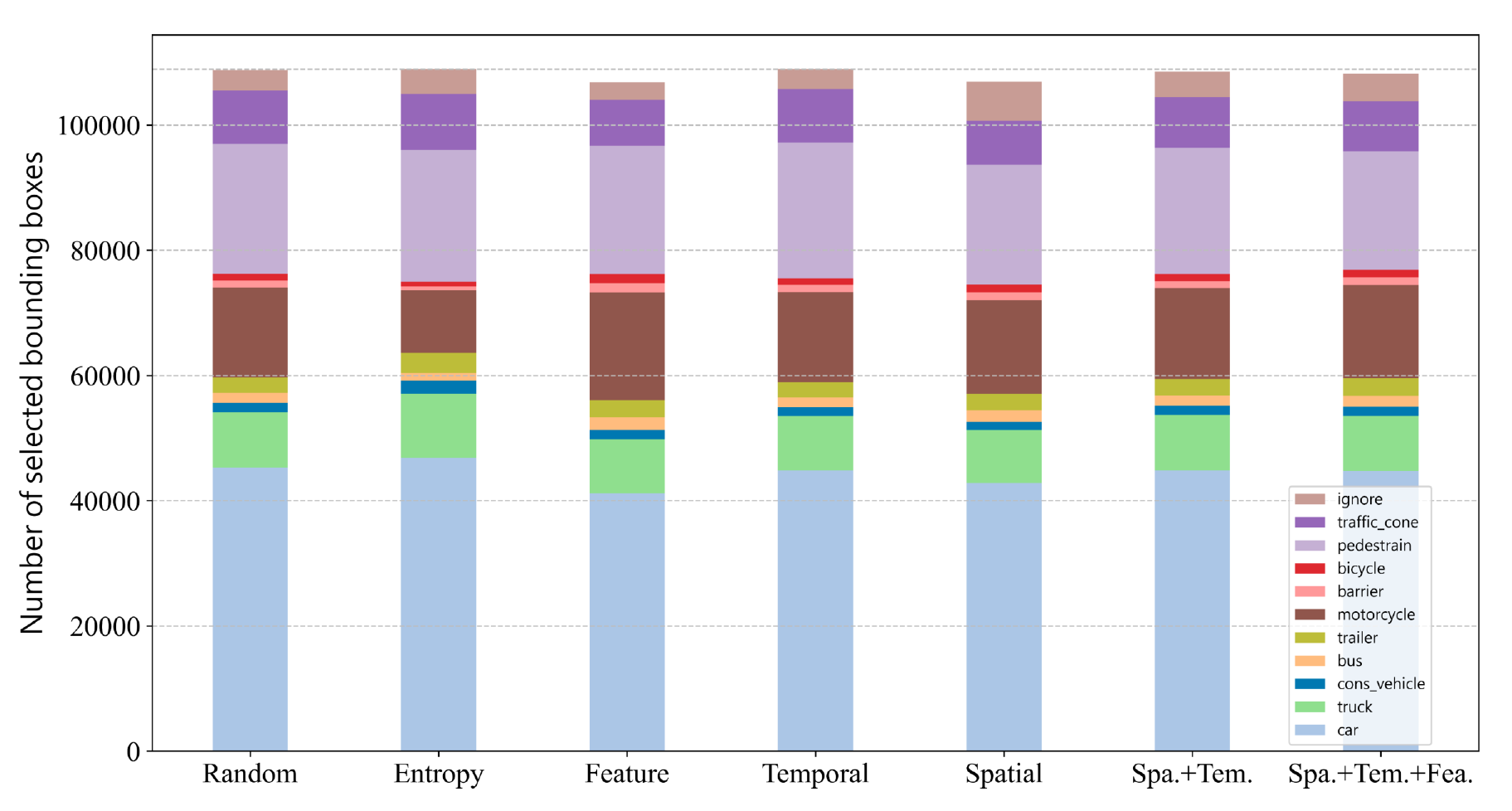}
    \caption{Distribution of selected bounding boxes over categories at budget 4800.}
    \label{FigBar}
\end{figure}

\begin{figure}[!tb]
    \centering
    \includegraphics[width=1.0\linewidth]{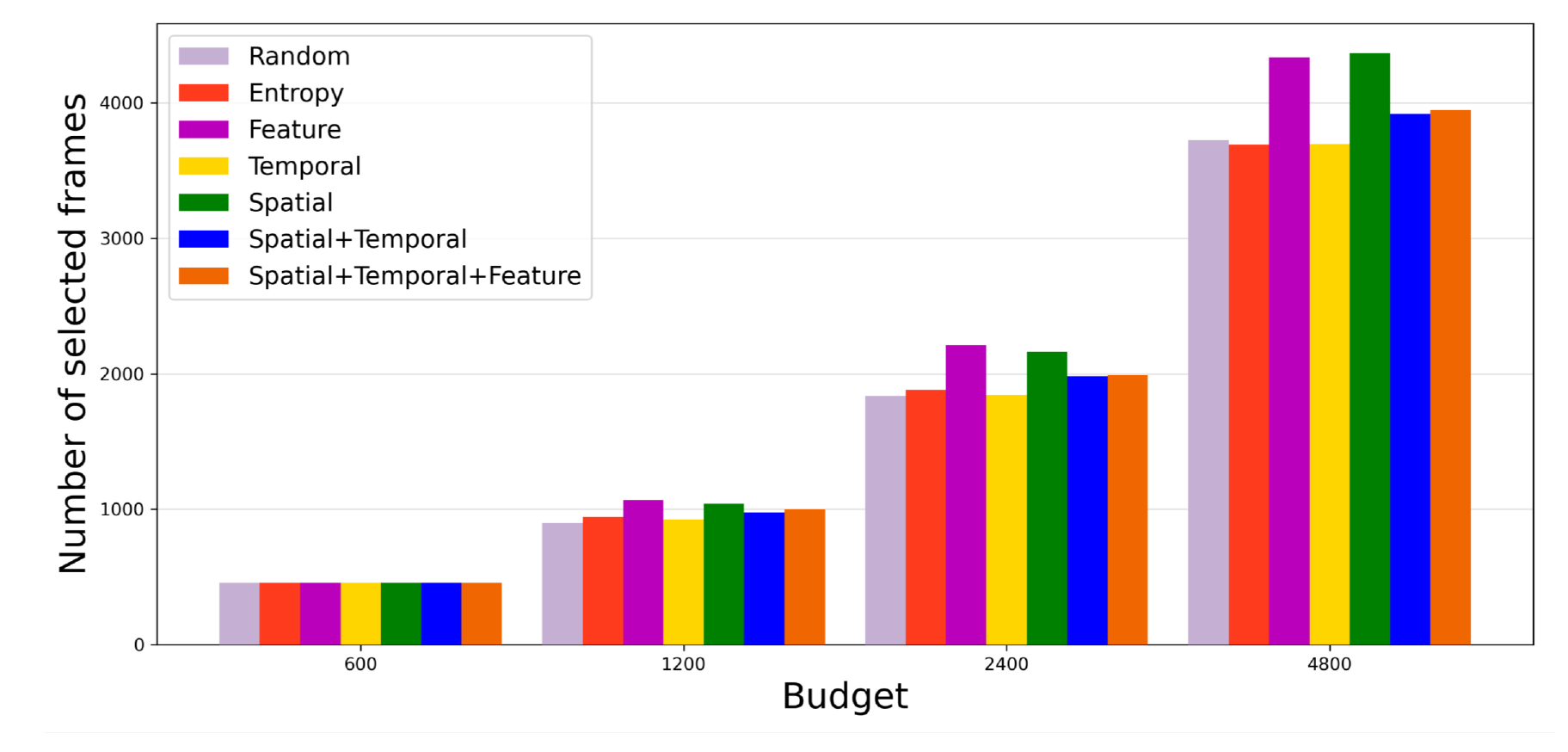}
    \caption{Number of frames selected by different strategies at fixed budgets.}
    \label{FigScenes}
\end{figure}

\subsection{{Qualitative Results}}

\begin{figure*}[!htb]
    \centering
    \includegraphics[width=0.95\linewidth]{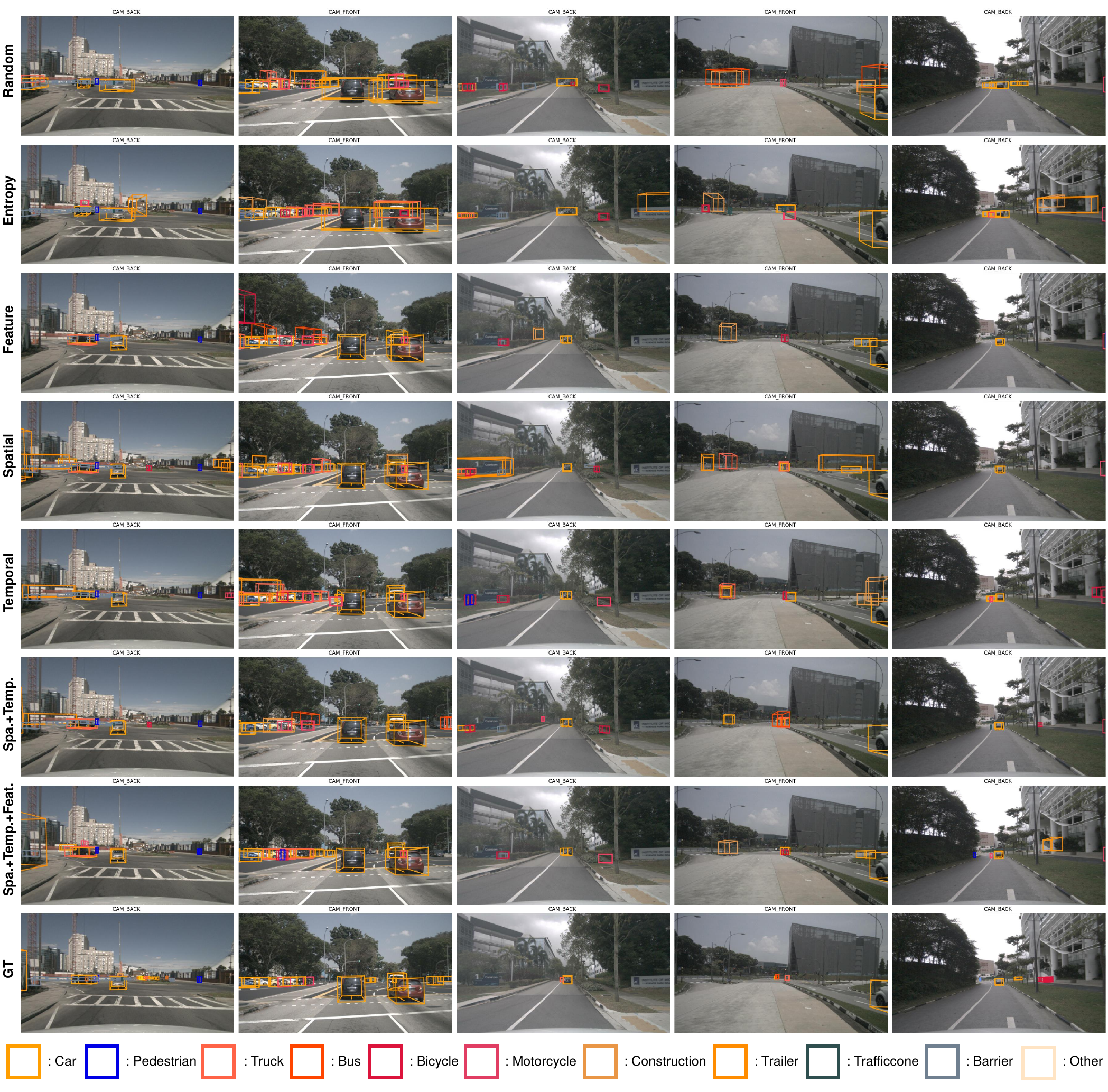}
    \caption{{Qualitative comparison of 3D object detection results with the same annotation budget on RGB view. Color of bounding box indicates the semantic category.}}
    \label{Figcam_det}
\end{figure*}

\begin{figure*}[!htb]
    \centering
    \includegraphics[width=0.95\linewidth]{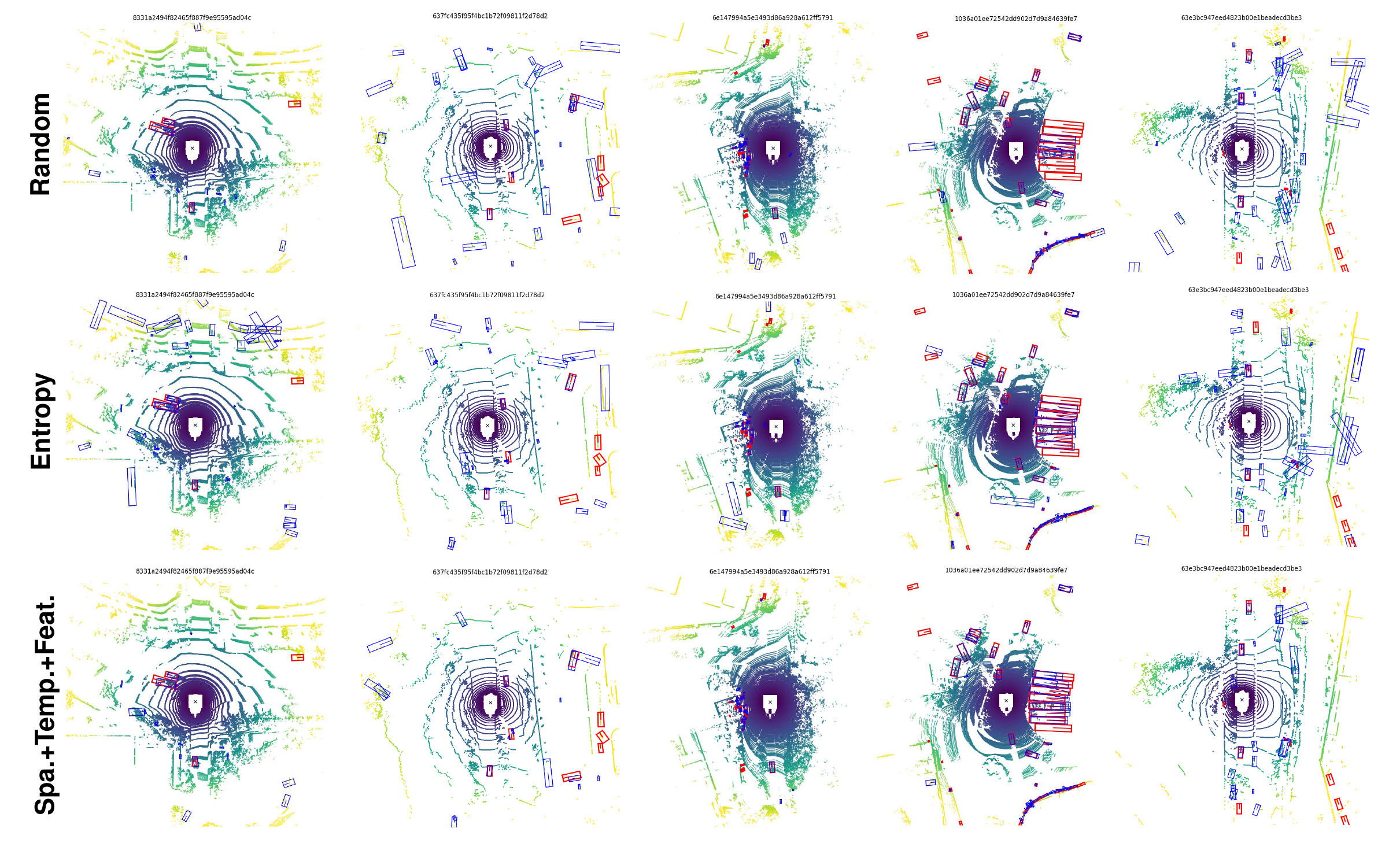}
    \caption{{Qualitative comparison of 3D object detection results with the same annotation budget on LiDAR point cloud view. We use red and blue boxes to indicate ground-truth annotations and detection results respectively.}}
    \label{Figlidar_det}
\end{figure*}

{In this section, we further present qualitative results to compare the effectiveness of different active selection strategies. Specifically, we visualize the detection results by the model trained with different active selection methods and ground-truth objects on both RGB view, in Fig.~\ref{Figcam_det} and LiDAR point cloud view, in Fig.~\ref{Figlidar_det}. All results are obtained with a budget of 4800. From both figures, We observe that incorporating all three diversity terms yields very competitive object detection results. In comparison, entropy and feature based approaches are overly focused on selecting less confident or diverse samples for training. This will compromise the generalization of trained model, as seen from the high false positive detections in Fig.~\ref{Figlidar_det}. }

\section{Discussion}
In this section, we provide more in-depth analysis and insights into the effectiveness of active learning on object detection for autonomous driving.

\subsection{{Alternative Object Detection Backbone}}

{In this section, we further evaluate the effectiveness of the proposed active learning strategy with BEVFusion backbone~\cite{liu2023bevfusion}. This backbone unifies multi-modal features in the shared bird's-eye view (BEV) representation space, which nicely preserves both geometric and semantic information. We present the comparisons of different active learning strategies with BEVFusion backbone in Tab.~\ref{TableCompDiveristy}. We observe that the overall performance with BEVFusion is improved from VoxelNet due to exploiting additional information. More importantly, our active learning strategy (Spa.+Temp. \& Spa.+Temp.+Feat.) still outperforms other simple active selection strategies with a clear margin.}

% \begin{table*}[!htb]
% \centering
%     \setlength\tabcolsep{6pt} % default value: 6pt
%      \caption{{Comparing different diversity terms for active learning with \textbf{BEVFusion} backbone. \textbf{Spatial}, \textbf{Temporal} and \textbf{Features} are shortened as \textbf{Spa.}, \textbf{Temp.} and \textbf{Feat.} respectively. Feature term in \textbf{Spa.+Temp.+Feat.} is enabled at budget 1200. All numbers are in (\%).}}\label{tab:BEVFusion}
%     \resizebox{0.99\linewidth}{!}{%
%     \begin{tabular}{c|ccccccc}
%     \hline
%      Budget & Random & Entropy~\cite{holub2008entropy} & Feature~\cite{sener2017active} & Spatial~\cite{cai2021exploring} & Temporal & Spa.+Temp. & Spa.+Temp.+Feat. \\
%     \hline
%     600 & 31.29$\pm$1.56 & 31.29$\pm$1.56 & 31.29$\pm$1.56 & 31.29$\pm$1.56 & 31.29$\pm$1.56 & 31.29$\pm$1.56 & 31.29$\pm$1.56 \\
%     1200 & 41.80$\pm$0.42 & 40.65$\pm$1.54 & 43.12$\pm$0.57 & 42.74$\pm$1.74 & 42.42$\pm$0.60 & \textbf{44.06$\pm$1.28} & 43.71$\pm$0.95 \\
%     2400 & 50.13$\pm$0.60 & 47.27$\pm$1.01 & 50.71$\pm$0.42 & \textbf{52.66$\pm$0.32} & 51.27$\pm$0.72 & 52.66$\pm$1.59 & 52.43$\pm$0.75 \\
%     4800 & 56.76$\pm$1.32 & 52.03$\pm$0.92 & 56.98$\pm$1.66 & 57.92$\pm$1.13 & 56.92$\pm$1.30 & 58.12$\pm$1.16 & \textbf{58.24$\pm$1.08} \\
%     \hline
%     \end{tabular}
%     }
% \end{table*}

\subsection{Selected Sample Distribution}
{To cast more insight into why diversity-based AL works better than uncertainty-based AL, we visualize %the distribution of frames selected by different acquisition functions in Fig.~\ref{FigHead}, 
the category distribution of selected bounding boxes in Fig.~\ref{FigBar}, and the number of selected frames in Fig.~\ref{FigScenes}. It can be seen that Entropy selects the most number of bounding boxes and the least number of frames. This suggests that relying on model uncertainty tends to concentrate the annotation budget in regions with high object density, and these regions typically correspond to busy streets where the number of objects within each scene is high. On the contrary, \textbf{Spatial} selects the most number of frames and the least number of bounding boxes, and the selected frames are distributed uniformly over all regions. The distribution of samples selected by \textbf{Feature} and \textbf{Temporal} lie in the middle, and the diversity terms complement each other to select an informative subset of samples to label.}

\begin{figure*}[!tb]
    \centering
    \includegraphics[width=1.0\linewidth]{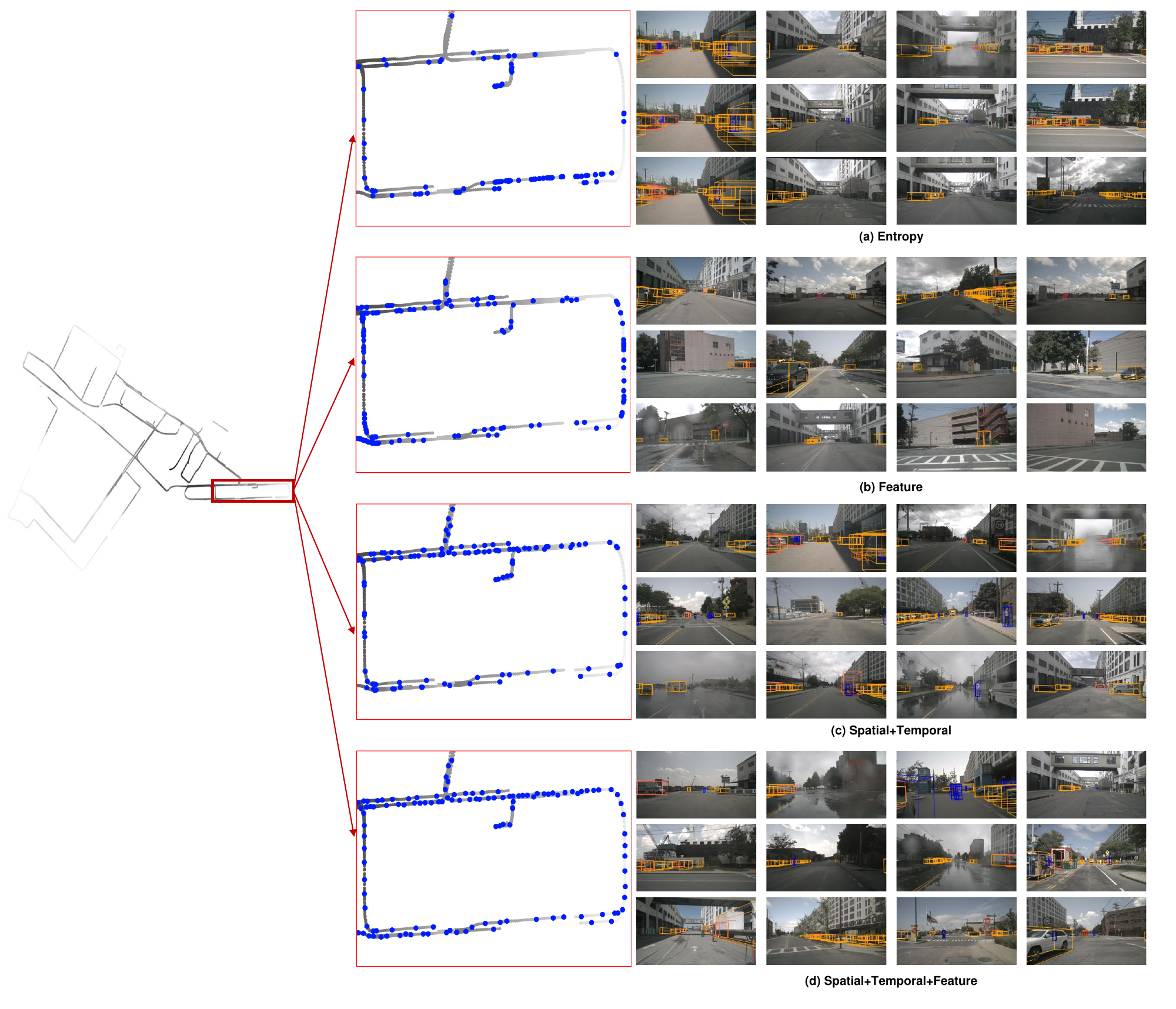}
    \caption{{A visualization of samples selected by different acquisition functions. Each {\textbullet} indicates selected frames on the driving trajectory by different active learning strategies.}}
    \label{FigSampling}
\end{figure*}

{We further qualitatively visualize the selected frames by different selection strategies at one segment of the road in Fig.~\ref{FigSampling}. For each strategy, we randomly select 12 frames and visualize the corresponding RGB views taken by the front camera with 3D boxes overlaid. We draw the following conclusions from the illustration. First, active learning algorithm considering \textbf{Uncertainty}~
alone tends to select redundant frames, e.g. many repetitive frames are selected in Fig.~\ref{FigSampling}~(a). Second, \textbf{Feature} diversity criteria tends to select more frames with fast changing background, e.g. many frames at bending corners are selected as shown in Fig.~\ref{FigSampling}~(b). In comparison, combining \textbf{Spatial} and \textbf{Temporal} diversity allows us to select frames with very diverse scenes, as shown in Fig.~\ref{FigSampling}~(c-d) which are demonstrated to produce better object detection performance.}

% It can be seen that \textbf{Uncertainty} tends to select highly similar samples. \textbf{Feature} is able to select more diverse samples but there is still some redundancy. \textbf{Spatial+Temporal} can complement \textbf{Feature} and further improve the diversity of selected samples. 

\begin{comment}
We show the number of the selected frames in Fig. \ref{FigScenes} and the statics of annotated bounding boxes in Fig. \ref{FigBar} for different AL methods. Due to the fact that the vehicle velocity will be affected by the busyness of the road, the sample densities in different regions are different, as stated before.
Given a fixed budget, \textbf{Entropy} selects the least frames and annotates the most bounding boxes. Such phenomenon demonstrates that the complex regions are challenging for the detector, which is intuitive. However, combining with the sampling results illustrated in Fig. \ref{FigHead}(a), \textbf{Entropy} redundantly selects samples from the complex regions, which harms the learning of the object detector, as illustrated in Fig. \ref{FigUncertainty}. We find that using spatial prior can effectively avoid redundant sampling in the complex regions, even \textbf{Spatial} selects the most frames and annotates the least bounding boxes. However, the organizers of AV datasets may plan the vehicle to drive in the same complex regions several times vary in traffic and weather conditions. Obviously, merely considering the spatial prior violates the organizers' intentions of their planning, which leads to poor description of the complex regions. Thus, simultaneously considering the spatial and temporal priors can achieve high-quality coverage of complex regions without redundant sampling. Table \ref{TableCompDiveristy} and Table \ref{TableStartUp} verify our hypothesis.
\end{comment}

\subsection{Initial Batch Selection}\label{SubSecStartup}
Typically, the AL acquisition function is contingent on the model prediction. For example, uncertainty-based AL depends on the model prediction score, and feature diversity-based AL utilizes the model for feature extraction. At the early stage of the AL cycles, due to data scarcity and model instability, AL strategies may fail. This is referred to as the cold-start problem of AL \cite{yuan2020cold}. One of the advantages of our proposed spatial and temporal diversity is that it does not rely on model prediction, and thus can be utilized to select betters samples for initializing the AL cycles.

In Section~\ref{sec:exp_results}, we fix the initial batch (selected by \textbf{Random}) for all methods for fair comparison. \Rev{In Table~\ref{tab:start_up}, we provide additional results, with warm start, where the initial batch is selected by \textbf{Spatial+Temporal}. We observe that warm start benefits more on the method with feature diversity~(Spatial+Temporal+Feature) by achieving better results in the long run. While Spatial+Temporal does not benefit from warm start because a better trained model has no impact of the selection of samples to annotate.} 

\begin{table}[!htb]
    \centering
    % \scalebox{0.75}{
        \setlength\tabcolsep{2pt} % default value: 6pt
        \caption{Comparing the effect of selecting the initial batch by \textbf{Random} (cold) or \textbf{Spatial+Temporal} (warm) with VoxelNet backbone. \Rev{All numbers are in (\%).}}
    \resizebox{1\linewidth}{!}{%
    \begin{tabular}{c|cccc}
    \hline
     & 600 & 1200 & 2400 & 4800 \\
    \hline
    Random              & 22.27$\pm$0.63 & 28.76$\pm$0.01 & 36.16$\pm$0.43 & 42.07$\pm$0.04 \\
    \hline
    Spa.+Temp.~(cold)    & 22.27$\pm$0.63 & \textbf{30.90$\pm$0.60} & \textbf{38.10$\pm$0.95} & \textbf{44.20$\pm$0.38} \\
    Spa.+Temp.~(warm)    & \textbf{23.87$\pm$0.01} & 30.56$\pm$0.26 & 37.96$\pm$0.27 & 43.94$\pm$0.24 \\
    \hline
    Spa.+Temp.+Feat.~(cold)  & 22.27$\pm$0.63 & \textbf{30.75$\pm$0.48} & 38.06$\pm$0.07 & {44.49$\pm$0.35} \\   
    Spa.+Temp.+Feat.~(warm)  & \textbf{23.87$\pm$0.01} & 30.56$\pm$0.27 & \textbf{38.30$\pm$0.33} & \textbf{45.38$\pm$0.21}  \\
    \hline
    \end{tabular}
    }
    \label{tab:start_up}
\end{table}

\subsection{Influence of Cost Measurement}\label{sec:influence_cost}
In previous experiments, we use the default annotation cost as 0.12 per frame ($c_f$) and the 0.04 per 3D bounding box ($c_b$). These values reflect the market rates charged by big data annotation companies in the US~\footnote{We would like to thank Dr. Holger Caesar for the invaluable discussion on realistic budgets for annotating autonomous driving data.}. Nevertheless, the ratio $c_f/c_b$ may vary among companies and may affect the evaluation results. Therefore, to verify the robustness of the proposed method under more diverse annotation cost schemes, we experiment with two additional cost schemes. First, we assume vendors only charge based on the number of bounding boxes, this gives rise to a cost scheme $c_f=0,c_b=0.04$. We further assume vendors only charge based on the number of annotated frames which gives rise to a cost scheme $c_f=0.6,c_b=0$. We present the active learning results for the above two cost schemes in Table~\ref{tab:influ_cost}. We make an observation from the results that regardless of annotation cost scheme, combining spatial, temporal and feature diversity consistently gives competitive performance compared with considering individual diversity terms and random selection baseline. The advantage of combining all three terms is particularly effective if cost is only counted on the number of bounding boxes or frames separately.  %It can be seen that how the cost is measured can significantly affect the relative ranking or the gap of different AL strategies. For example, when only frame annotation is charged, \textbf{Random} emerges as a strong baseline, and the gap between \textbf{Entropy} and other methods is greatly reduced. This highlights the importance of using realistic annotation cost measurement to properly evaluate different methods.

\begin{table*}[!htb]
    \centering
    % \scalebox{0.75}{
    \setlength\tabcolsep{3pt} % default value: 6pt
    \caption{Evaluation of sensitivity to cost measurements. All methods select the first batch by \textbf{Random}. \textcolor{red}{Red}, \textcolor{blue}{blue} and \textcolor{green}{green} colors indicate the best, second best and the third best results respectively. \Rev{All numbers are in (\%).}} \label{tab:influ_cost}
    \resizebox{1\linewidth}{!}{%
    \begin{tabular}{c|ccc|ccc|ccc}
    \hline
     & 600 & 1200 & 2400 & 600 & 1200 & 2400 & 600 & 1200 & 2400 \\
    \hline
     & \multicolumn{3}{c|}{$c_f=0.12, c_b=0.04$} & \multicolumn{3}{c}{$c_f=0.0, c_b=0.04$} & \multicolumn{3}{|c}{$c_f=0.6, c_b=0.0$} \\
    \hline
    Random              & {22.27$\pm$0.63} & 28.76$\pm$0.01 & 36.16$\pm$0.43 & {22.96$\pm$0.62} & 29.93$\pm$0.63 & 37.06$\pm$0.42 & {29.49$\pm$0.52} & 37.18$\pm$0.41 & 42.40$\pm$0.35 \\
    Entropy~\cite{holub2008entropy}             & {22.27$\pm$0.63} & 27.67$\pm$0.49 & 32.30$\pm$0.07 & {22.96$\pm$0.62} & 28.75$\pm$0.28 & 33.93$\pm$0.55 & {29.49$\pm$0.52} & 34.52$\pm$0.12 & 39.39$\pm$0.51 \\
    Feature~\cite{sener2017active}             & {22.27$\pm$0.63} & 29.77$\pm$0.55 & 35.84$\pm$0.35 & {22.96$\pm$0.62} & 30.35$\pm$0.23 & 37.15$\pm$0.36 & {29.49$\pm$0.52} & 35.51$\pm$0.38 & 41.48$\pm$0.36 \\
    Spatial~\cite{cai2021exploring}             & {22.27$\pm$0.63} & \textcolor{green}{30.25$\pm$0.68} & \textcolor{red}{38.12$\pm$0.13} & {22.96$\pm$0.62} & \textcolor{blue}{31.61$\pm$0.07} & \textcolor{green}{38.81$\pm$0.34} & {29.49$\pm$0.52} & 36.49$\pm$0.26 & 43.08$\pm$0.34 \\
    Temporal~(Ours)            & {22.27$\pm$0.63} & 29.81$\pm$0.05 & 37.34$\pm$0.23 & {22.96$\pm$0.62} & 30.90$\pm$0.41 & 37.98$\pm$0.18 & {29.49$\pm$0.52} & \textcolor{green}{37.59$\pm$0.14} & \textcolor{green}{43.15$\pm$0.45} \\
    Spatial+Temp.~(Ours)    & {22.27$\pm$0.63} & \textcolor{red}{30.90$\pm$0.60} & \textcolor{blue}{38.10$\pm$0.95} & {22.96$\pm$0.62} & \textcolor{green}{31.07$\pm$0.24} & \textcolor{blue}{39.23$\pm$0.23} & {29.49$\pm$0.52} & \textcolor{blue}{37.74$\pm$0.17} & \textcolor{blue}{44.26$\pm$0.08}\\
    Spa.+Temp.+Feat.~(Ours)    & {22.27$\pm$0.63} & \textcolor{blue}{30.75$\pm$0.48} & \textcolor{green}{38.06$\pm$0.07} &{22.96$\pm$0.62} &\textcolor{red}{31.67$\pm$0.39}  &\textcolor{red}{39.33$\pm$0.43} & {29.49$\pm$0.52}& \textcolor{red}{37.95$\pm$0.21} &  \textcolor{red}{44.67$\pm$0.21}\\
    \hline
    \end{tabular}
    }
\end{table*}

\subsection{Alternative Methods for Diversity Aggregation and Normalization} \label{SubSecAggregate}
We investigate alternative aggregation and normalization methods for combing various diversity terms. The aggregated value can be obtained by computing the summation, minimum or maximum of the individual terms. The results with alternative aggregation methods are presented in Table~\ref{TableAggregaed}. Experiments are evaluated with \textbf{Spatial+Temporal}. It can be seen that summation performs the best. We also experiment with linear scaling for normalization, where the distance value is normalized to [0,1] by dividing by the maximum value of the distance matrix. The results are presented in Table~\ref{TableNormalized}. It can be seen that RBF performs better than linear normalization. % \cll{should provide the full name for RBF somewhere in the paper. Also, what is gamma used in the RBF kernel? Ablation should be provided.} 

%As stated in Section \ref{SubSecAcquisition}, we convert all distance matrices into affinity matrices through RBF kernel to realization normalization. Formulating a reliable optimization objective is the key of diversity-based AL methods. Giving all affinity matrices, the optimization objective can be aggregated by summation, minimum or maximum. Table \ref{TableAggregaed} shows that summation performs the best, which means the summation can realize the best aggregation of spatial and temporal priors. Otherwise, the comparison between L1 and Sum verifies that the RBF kernel realizes the normalization better.

\begin{table}[!htb]
    \centering
    \setlength\tabcolsep{3pt} % default value: 6pt
        \caption{Ablation study on the aggregation methods for combining the diversity terms. \Rev{All numbers are in (\%).}}
    \resizebox{0.95\linewidth}{!}{%
    \begin{tabular}{c|cccc}
    \hline
    Aggregation & 600 & 1200 & 2400 & 4800 \\
    \hline
    % \hline
    Max & \textbf{22.27$\pm$0.63}  & 28.77$\pm$0.53 & 37.56$\pm$0.37 & 43.58$\pm$0.41  \\
    Min & \textbf{22.27$\pm$0.63}  & 30.72$\pm$0.56 & \textbf{38.63$\pm$0.30} & 43.87$\pm$0.36\\
    Sum & \textbf{22.27$\pm$0.63} & \textbf{30.90$\pm$0.60} & 38.10$\pm$0.95 & \textbf{44.20$\pm$0.38} \\
    \hline
    \end{tabular}
    }
    % \vspace{-0.5cm}
    %\caption{Ablation study on the aggregation of the spatial and temporal affinity matrices. L1 represents that the affinity matrices obtained by dividing the distance matrices by their maximums and using summation to aggregate the affinity matrices. Others obtain the affinity matrices through the RBF kernel.}
    \label{TableAggregaed}
\end{table}

\begin{table}[!htb]
    \centering
        \caption{Ablation study on the normalization methods for combining the diversity terms.}\label{TableNormalized}
    \resizebox{0.9\linewidth}{!}{%
    \begin{tabular}{c|cccc}
    \hline
    Normalization  & 1200 & 2400 & 4800 \\
    \hline
    Linear   & 29.95$\pm$0.22 & 37.60$\pm$0.48 & 43.76$\pm$0.10 \\
    RBF   & \textbf{30.90$\pm$0.60} & \textbf{38.10$\pm$0.95} & \textbf{44.20$\pm$0.38} \\
    \hline
    \end{tabular}
    }
    % \vspace{-0.5cm}
\end{table}

%%%%%%%%%%%%%%%%%%%%%%%

\subsection{Analysis on Spatial Distance Measurement}
For spatial diversity term, we use the manifold distance to measure the spatial distance between two samples. This is motivated by the intuition that given pairs of samples with the same Euclidean distance, the pair collected along the same road may be more similar, i.e. closer, to each other compared to those collected from different roads. We also experimented with Euclidean distance and the results are presented in Table \ref{TableSpatial}. It can be seen that manifold distance slightly outperforms Euclidean distance.

\begin{table}[!htb]
    \centering
    \caption{Comparing manifold and Euclidean distance for defining spatial diversity. \label{TableSpatial}}
    \setlength\tabcolsep{4pt} % default value: 6pt
    \resizebox{0.95\linewidth}{!}{%
    \begin{tabular}{c|cccc}
    \hline
     & 600 & 1200 & 2400 & 4800 \\
    \hline
    Euclidean   & {22.27$\pm$0.63} & 29.63$\pm$0.47 & 37.81$\pm$0.54 & \textbf{43.93$\pm$0.08} \\
    Manifold    & {22.27$\pm$0.63} & \textbf{30.25$\pm$0.68} & \textbf{38.12$\pm$0.13} & 43.84$\pm$0.22 \\
    \hline
    \end{tabular}
    }
\end{table}

\subsection{Evaluation of Hyper-Parameter}

\Rev{We further evaluate the stability of proposed method w.r.t. some hyper-parameters. In particular, $\lambda_s, \lambda_t, \lambda_f$ in Eq.~\ref{EqDagree} determine the distance metric. We evaluate how the ratio between $\lambda_s$ and $\lambda_t$ could affect the active learning performance with \textbf{Spatial+Temporal}. As shown in Tab.~\ref{TableLambdaT}, the performance is relatively stable around $\lambda_t=0.5$.}

\begin{table}[!htb]
    \centering
    \caption{\Rev{Impact of the hyper-parameters $\lambda_s$ and $\lambda_t$ on active learning performance. All numbers are in (\%).}}\label{TableLambdaT}
    \resizebox{0.95\linewidth}{!}{%
    \begin{tabular}{c|ccccc}
    \hline
    $\lambda_t$ & 0 & 0.5 & 1 & 2 \\
    \hline
    1200 &  30.25$\pm$0.68 & 30.53$\pm$0.71 & \textbf{30.90$\pm$0.60} & 30.76$\pm$0.76\\
    2400 &  38.12$\pm$0.13 & \textbf{38.47$\pm$0.46} & 38.10$\pm$0.95 & 37.97$\pm$0.31\\
    4800 & 43.84$\pm$0.22 & \textbf{44.75$\pm$0.09} & 44.20$\pm$0.38 & 44.24$\pm$0.07\\
    \hline
    \end{tabular}
    }
\end{table}

\Rev{To study the effect of $\lambda_f$ on \textbf{Spatial+Temporal+Feature}, we vary the value from 0 to 5 and the results are presented in Table \ref{TableLambdaF}. We validate that the performance is stable around $\lambda_f=1$.}

\begin{table}[!htb]
    \centering
    \caption{\Rev{Effect of $\lambda_f$ on \textbf{Spatial+Temporal+Feature}. We turn on the feature term after budget 1200.} \label{TableLambdaF}}
    \setlength\tabcolsep{3pt} % default value: 6pt
    \resizebox{1\linewidth}{!}{%
    \begin{tabular}{c|ccccc}
    \hline
    $\lambda_f$ & 0 & 0.1 & 1 & 2 & 5 \\
    \hline
    1200 & \textbf{30.90$\pm$0.60} & 30.62$\pm$0.25 & 30.75$\pm$0.48 & 30.18$\pm$0.35 & 30.53$\pm$0.20 \\
    2400 & 38.10$\pm$0.95 & \textbf{38.13$\pm$0.61} & 38.06$\pm$0.07 & 38.01$\pm$0.24 & 37.98$\pm$0.31 \\
    4800 & 44.20$\pm$0.38 & 44.37$\pm$0.38 & 44.49$\pm$0.35 & \textbf{44.61$\pm$0.11} & 44.46$\pm$0.58 \\
    \hline
    \end{tabular}
    % \begin{tabular}{c|cccc}
    % \hline
    % $\lambda_f$ & 1200 & 2400 & 4800 \\
    % \hline
    % 0   & \textbf{32.03} & \textbf{39.51} & 44.85 \\
    % 0.1 & \textbf{32.03} & 38.87 & 44.84 \\
    % 1   & \textbf{32.03} & 39.23 & \textbf{45.02} \\
    % 2   & \textbf{32.03} & 39.31 & 44.95 \\
    % 5   & \textbf{32.03} & 38.50 & 44.49 \\
    % \hline
    % \end{tabular}
    }
\end{table}

\subsection{Analysis on Feature Diversity Term}
We combine the spatial, temporal and feature diversity terms as Eq.\ref{EqDagree}.
Compared to spatial and temporal terms, the feature diversity term is contingent on the model prediction. At the early stage of the AL cycles, the model may be unreliable and affect the effectiveness of the extracted features. It may be beneficial to enable feature diversity only at later AL cycles. We investigate the optimal budget size to turn on feature diversity for active selection. %and the effect of $\lambda_f$.
we experiment with \textbf{Spatial+Temporal+Feature} where the feature term is only enabled after certain amount of budgets. We fix $\lambda_f$ to be 1 when the feature term is enabled. The results reported in Table \ref{TablestartSize} show that turning on feature diversity after 1200 cost achieves slightly better results. The trade-off is that, if the feature term is enabled too early, the features are not reliable and can harm the performance; on the other hand, if enabled too late, the information from current model is under-utilized, which may also be sub-optimal.

\begin{table}[!htb]
    \centering
     \caption{Investigating the optimal budget size to enable feature diversity for \textbf{Spatial+Temporal+Feature}. \textbf{Baseline} represents \textbf{Spatial+Temporal} where the feature diversity is not considered. \label{TablestartSize}}
     \setlength\tabcolsep{4pt} % default value: 6pt
    \resizebox{1\linewidth}{!}{%
    \begin{tabular}{c|cccc}
    \hline
    start size & 600 & 1200 & 2400 & 4800 \\
    \hline
    Baseline    & \textbf{22.27$\pm$0.63} & \textbf{30.90$\pm$0.60} & {38.10$\pm$0.95} & 44.20$\pm$0.38 \\
    600         & \textbf{22.27$\pm$0.63} & 30.75$\pm$0.48 & 38.06$\pm$0.07 & \textbf{44.49$\pm$0.35} \\
    1200        & \textbf{22.27$\pm$0.63} & \textbf{30.90$\pm$0.60} & \textbf{38.19$\pm$0.43} & 44.13$\pm$0.13 \\
    2400        & \textbf{22.27$\pm$0.63} & \textbf{30.90$\pm$0.60} & 38.10$\pm$0.95 & 44.21$\pm$0.07 \\
    \hline
    \end{tabular}
    }
\end{table}

% \subsection{Influence of Cost Measurement at Larger Budget}
% We add experiments at budget 4800 for the three annotation budget settings used in the paper, i.e. $c_f=0,c_b=0.04/c_f=0.12,c_b=0.04/c_f=0.6,c_b=0.04$. The results are shown in Fig.~\ref{FigCost}. The observation is consistent with previous studies, the measurement of annotation cost, e.g. considering different costs for each frame and bounding box, will have an impact on the overall performance. Given a fixed budget 4800, the most realistic way towards annotation counting, i.e. emphasizing both frame and bounding-box annotation cost, will result in lower performance. But regardless of the cost measurements, our approach (Spatial+Temporal) always outperforms the alternative active learning methods.

% \begin{figure}[!tb]
%     \centering
%     \includegraphics[width=1.0\linewidth]{images/cost_influence_new.jpg}
%     \caption{Influence of cost measurement at budget 2400. Dash lines represent the best performing method under different cost measurements (represented by different colors). \textbf{0.0,0.04}: $c_f=0,c_b=0.04$; \textbf{0.12,0.04}: $c_f=0.12,c_b=0.04$; \textbf{0.6,0.0}: $c_f=0.6,c_b=0.0$.}
%     \label{FigCost}
% \end{figure}

%%%%%%%%%%%%%%%%%%%%%%%

\subsection{Limitations}
Our proposed method leverages the data collected from the GPS/IMU system to enforce diversity in spatial space. In practice, the quality of the collected data can be affected by device outages, especially for vehicles operating in dense urban areas. Some datasets, e.g. nuScenes, employed additional techniques, e.g. HD map, to guarantee accurate localization, but other datasets, e.g. KITTI, did not. The localization errors may affect the effectiveness of the proposed spatial diversity term. The proposed temporal diversity term and feature diversity term complement spatial diversity and help to alleviate this limitation. \Rev{We also point out that the performance gap between methods considering different diversity terms seems marginal. Nevertheless, as the scale of dataset increases, the cost saved from selecting informative samples could be substantial. }

\section{Conclusion}

In this paper, we introduced a novel diversity-based active learning (AL) approach for 3D object detection, focusing on spatial and temporal diversity. Our experiments on the nuScenes dataset show that our method outperforms all existing active learning baselines and active learning methods specifically tailored for 2D object detection tasks. We also proposed a realistic annotation cost measurement, which significantly impacts the ranking of AL strategies. Additionally, our approach addresses the cold-start problem by effectively selecting an initial batch that enhances both early and long-term performance. Our work pioneers AL in 3D object detection for autonomous driving, offering valuable insights and practical guidelines for future research.

\ifCLASSOPTIONcaptionsoff
  \newpage
\fi



%\bibliographystyle{IEEEtran}
\bibliographystyle{IEEEtran}

\bibliography{mybibfile}

\begin{thebibliography}{10}
\providecommand{\url}[1]{#1}
\csname url@samestyle\endcsname
\providecommand{\newblock}{\relax}
\providecommand{\bibinfo}[2]{#2}
\providecommand{\BIBentrySTDinterwordspacing}{\spaceskip=0pt\relax}
\providecommand{\BIBentryALTinterwordstretchfactor}{4}
\providecommand{\BIBentryALTinterwordspacing}{\spaceskip=\fontdimen2\font plus
\BIBentryALTinterwordstretchfactor\fontdimen3\font minus
  \fontdimen4\font\relax}
\providecommand{\BIBforeignlanguage}[2]{{%
\expandafter\ifx\csname l@#1\endcsname\relax
\typeout{** WARNING: IEEEtran.bst: No hyphenation pattern has been}%
\typeout{** loaded for the language `#1'. Using the pattern for}%
\typeout{** the default language instead.}%
\else
\language=\csname l@#1\endcsname
\fi
#2}}
\providecommand{\BIBdecl}{\relax}
\BIBdecl

\bibitem{yang20203dssd}
Z.~Yang, Y.~Sun, S.~Liu, and J.~Jia, ``3dssd: Point-based 3d single stage
  object detector,'' in \emph{Proceedings of the IEEE/CVF conference on
  computer vision and pattern recognition}, 2020, pp. 11\,040--11\,048.

\bibitem{zhu2019class}
B.~Zhu, Z.~Jiang, X.~Zhou, Z.~Li, and G.~Yu, ``Class-balanced grouping and
  sampling for point cloud 3d object detection,'' \emph{arXiv preprint
  arXiv:1908.09492}, 2019.

\bibitem{yin2021center}
T.~Yin, X.~Zhou, and P.~Krahenbuhl, ``Center-based 3d object detection and
  tracking,'' in \emph{Proceedings of the IEEE/CVF Conference on Computer
  Vision and Pattern Recognition}, 2021, pp. 11\,784--11\,793.

\bibitem{qi2018frustum}
C.~R. Qi, W.~Liu, C.~Wu, H.~Su, and L.~J. Guibas, ``Frustum pointnets for 3d
  object detection from rgb-d data,'' in \emph{Proceedings of the IEEE
  conference on computer vision and pattern recognition}, 2018, pp. 918--927.

\bibitem{Shi_2019_CVPR}
S.~Shi, X.~Wang, and H.~Li, ``Pointrcnn: 3d object proposal generation and
  detection from point cloud,'' in \emph{The IEEE Conference on Computer Vision
  and Pattern Recognition (CVPR)}, June 2019.

\bibitem{Lang_2019_CVPR}
A.~H. Lang, S.~Vora, H.~Caesar, L.~Zhou, J.~Yang, and O.~Beijbom,
  ``Pointpillars: Fast encoders for object detection from point clouds,'' in
  \emph{Proceedings of the IEEE/CVF Conference on Computer Vision and Pattern
  Recognition (CVPR)}, June 2019.

\bibitem{yan2018second}
Y.~Yan, Y.~Mao, and B.~Li, ``Second: Sparsely embedded convolutional
  detection,'' \emph{Sensors}, vol.~18, no.~10, p. 3337, 2018.

\bibitem{roth2006margin}
D.~Roth and K.~Small, ``Margin-based active learning for structured output
  spaces,'' in \emph{European Conference on Machine Learning}.\hskip 1em plus
  0.5em minus 0.4em\relax Springer, 2006, pp. 413--424.

\bibitem{joshi2009multi}
A.~J. Joshi, F.~Porikli, and N.~Papanikolopoulos, ``Multi-class active learning
  for image classification,'' in \emph{2009 IEEE Conference on Computer Vision
  and Pattern Recognition}.\hskip 1em plus 0.5em minus 0.4em\relax IEEE, 2009,
  pp. 2372--2379.

\bibitem{yoo2019learning}
D.~Yoo and I.~S. Kweon, ``Learning loss for active learning,'' in
  \emph{Proceedings of the IEEE/CVF Conference on Computer Vision and Pattern
  Recognition}, 2019, pp. 93--102.

\bibitem{sener2017active}
O.~Sener and S.~Savarese, ``Active learning for convolutional neural networks:
  A core-set approach,'' in \emph{International Conference on Learning
  Representations}, 2018.

\bibitem{lin2017active}
L.~Lin, K.~Wang, D.~Meng, W.~Zuo, and L.~Zhang, ``Active self-paced learning
  for cost-effective and progressive face identification,'' \emph{IEEE
  transactions on pattern analysis and machine intelligence}, vol.~40, no.~1,
  pp. 7--19, 2017.

\bibitem{yang2015multi}
Y.~Yang, Z.~Ma, F.~Nie, X.~Chang, and A.~G. Hauptmann, ``Multi-class active
  learning by uncertainty sampling with diversity maximization,''
  \emph{International Journal of Computer Vision}, vol. 113, no.~2, pp.
  113--127, 2015.

\bibitem{elhamifar2013convex}
E.~Elhamifar, G.~Sapiro, A.~Yang, and S.~S. Sasrty, ``A convex optimization
  framework for active learning,'' in \emph{Proceedings of the IEEE
  International Conference on Computer Vision}, 2013, pp. 209--216.

\bibitem{guo2010active}
Y.~Guo, ``Active instance sampling via matrix partition.'' in \emph{NIPS},
  2010, pp. 802--810.

\bibitem{gal2015bayesian}
Y.~Gal and Z.~Ghahramani, ``Bayesian convolutional neural networks with
  bernoulli approximate variational inference,'' \emph{arXiv preprint
  arXiv:1506.02158}, 2015.

\bibitem{gal2017deep}
Y.~Gal, R.~Islam, and Z.~Ghahramani, ``Deep bayesian active learning with image
  data,'' in \emph{International Conference on Machine Learning}.\hskip 1em
  plus 0.5em minus 0.4em\relax PMLR, 2017, pp. 1183--1192.

\bibitem{li2017dropout}
Y.~Li and Y.~Gal, ``Dropout inference in bayesian neural networks with
  alpha-divergences,'' in \emph{International conference on machine
  learning}.\hskip 1em plus 0.5em minus 0.4em\relax PMLR, 2017, pp. 2052--2061.

\bibitem{aghdam2019active}
H.~H. Aghdam, A.~Gonzalez-Garcia, J.~v.~d. Weijer, and A.~M. L{\'o}pez,
  ``Active learning for deep detection neural networks,'' in \emph{Proceedings
  of the IEEE/CVF International Conference on Computer Vision}, 2019, pp.
  3672--3680.

\bibitem{roy2018deep}
S.~Roy, A.~Unmesh, and V.~P. Namboodiri, ``Deep active learning for object
  detection.'' in \emph{BMVC}, vol. 362, 2018, p.~91.

\bibitem{kao2018localization}
C.-C. Kao, T.-Y. Lee, P.~Sen, and M.-Y. Liu, ``Localization-aware active
  learning for object detection,'' in \emph{Asian Conference on Computer
  Vision}.\hskip 1em plus 0.5em minus 0.4em\relax Springer, 2018, pp. 506--522.

\bibitem{mackowiak2018cereals}
R.~Mackowiak, P.~Lenz, O.~Ghori, F.~Diego, O.~Lange, and C.~Rother,
  ``Cereals-cost-effective region-based active learning for semantic
  segmentation,'' \emph{arXiv preprint arXiv:1810.09726}, 2018.

\bibitem{cai2021exploring}
L.~Cai, X.~Xu, L.~Zhang, and C.-S. Foo, ``Exploring spatial diversity for
  region-based active learning,'' \emph{IEEE Transactions on Image Processing},
  2021.

\bibitem{cai2021revisiting}
L.~Cai, X.~Xu, J.~H. Liew, and C.~S. Foo, ``Revisiting superpixels for active
  learning in semantic segmentation with realistic annotation costs,'' in
  \emph{Proceedings of the IEEE/CVF Conference on Computer Vision and Pattern
  Recognition}, 2021, pp. 10\,988--10\,997.

\bibitem{feng2019deep}
D.~Feng, X.~Wei, L.~Rosenbaum, A.~Maki, and K.~Dietmayer, ``Deep active
  learning for efficient training of a lidar 3d object detector,'' in
  \emph{2019 IEEE Intelligent Vehicles Symposium (IV)}.\hskip 1em plus 0.5em
  minus 0.4em\relax IEEE, 2019, pp. 667--674.

\bibitem{yuan2021multiple}
T.~Yuan, F.~Wan, M.~Fu, J.~Liu, S.~Xu, X.~Ji, and Q.~Ye, ``Multiple instance
  active learning for object detection,'' in \emph{Proceedings of the IEEE/CVF
  Conference on Computer Vision and Pattern Recognition}, 2021, pp. 5330--5339.

\bibitem{desai2020towards}
S.~V. Desai and V.~N. Balasubramanian, ``Towards fine-grained sampling for
  active learning in object detection,'' in \emph{Proceedings of the IEEE/CVF
  Conference on Computer Vision and Pattern Recognition Workshops}, 2020, pp.
  924--925.

\bibitem{gao2020consistency}
M.~Gao, Z.~Zhang, G.~Yu, S.~{\"O}. Ar{\i}k, L.~S. Davis, and T.~Pfister,
  ``Consistency-based semi-supervised active learning: Towards minimizing
  labeling cost,'' in \emph{European Conference on Computer Vision}.\hskip 1em
  plus 0.5em minus 0.4em\relax Springer, 2020, pp. 510--526.

\bibitem{ren2020survey}
P.~Ren, Y.~Xiao, X.~Chang, P.-Y. Huang, Z.~Li, X.~Chen, and X.~Wang, ``A survey
  of deep active learning,'' \emph{arXiv preprint arXiv:2009.00236}, 2020.

\bibitem{houlsby2011bayesian}
N.~Houlsby, F.~Husz{\'a}r, Z.~Ghahramani, and M.~Lengyel, ``Bayesian active
  learning for classification and preference learning,'' \emph{arXiv preprint
  arXiv:1112.5745}, 2011.

\bibitem{fuchsgruber2024uncertainty}
D.~Fuchsgruber, T.~Wollschläger, B.~Charpentier, A.~Oroz, and S.~Günnemann,
  ``Uncertainty for active learning on graphs,'' 2024.

\bibitem{beluch2018power}
W.~H. Beluch, T.~Genewein, A.~N{\"u}rnberger, and J.~M. K{\"o}hler, ``The power
  of ensembles for active learning in image classification,'' in
  \emph{Proceedings of the IEEE Conference on Computer Vision and Pattern
  Recognition}, 2018, pp. 9368--9377.

\bibitem{safaei2024entropic}
B.~Safaei, V.~Vibashan, C.~M. de~Melo, and V.~M. Patel, ``Entropic open-set
  active learning,'' in \emph{Proceedings of the AAAI Conference on Artificial
  Intelligence}, vol.~38, no.~5, 2024, pp. 4686--4694.

\bibitem{li2024deep}
X.~Li, P.~Yang, Y.~Gu, X.~Zhan, T.~Wang, M.~Xu, and C.~Xu, ``Deep active
  learning with noise stability,'' in \emph{Proceedings of the AAAI Conference
  on Artificial Intelligence}, vol.~38, no.~12, 2024, pp. 13\,655--13\,663.

\bibitem{hasan2015context}
M.~Hasan and A.~K. Roy-Chowdhury, ``Context aware active learning of activity
  recognition models,'' in \emph{Proceedings of the IEEE International
  Conference on Computer Vision}, 2015, pp. 4543--4551.

\bibitem{tang2019self}
Y.-P. Tang and S.-J. Huang, ``Self-paced active learning: Query the right thing
  at the right time,'' in \emph{Proceedings of the AAAI conference on
  artificial intelligence}, vol.~33, no.~01, 2019, pp. 5117--5124.

\bibitem{sinha2019variational}
S.~Sinha, S.~Ebrahimi, and T.~Darrell, ``Variational adversarial active
  learning,'' in \emph{Proceedings of the IEEE/CVF International Conference on
  Computer Vision}, 2019, pp. 5972--5981.

\bibitem{shi2021label}
X.~Shi, X.~Xu, K.~Chen, L.~Cai, C.~S. Foo, and K.~Jia, ``Label-efficient point
  cloud semantic segmentation: An active learning approach,'' \emph{arXiv
  preprint arXiv:2101.06931}, 2021.

\bibitem{wu2021redal}
T.-H. Wu, Y.-C. Liu, Y.-K. Huang, H.-Y. Lee, H.-T. Su, P.-C. Huang, and W.~H.
  Hsu, ``Redal: Region-based and diversity-aware active learning for point
  cloud semantic segmentation,'' in \emph{Proceedings of the IEEE/CVF
  International Conference on Computer Vision}, 2021.

\bibitem{brust2018active}
C.-A. Brust, C.~K{\"a}ding, and J.~Denzler, ``Active learning for deep object
  detection,'' \emph{arXiv preprint arXiv:1809.09875}, 2018.

\bibitem{blad2023}
M.~Lyu, J.~Zhou, H.~Chen, Y.~Huang, D.~Yu, Y.~Li, Y.~Guo, Y.~Guo, L.~Xiang, and
  G.~Ding, ``Box-level active detection,'' in \emph{Proceedings of the IEEE/CVF
  Conference on Computer Vision and Pattern Recognition (CVPR)}, June 2023, pp.
  23\,766--23\,775.

\bibitem{park2023active}
Y.~Park, W.~Choi, S.~Kim, D.-J. Han, and J.~Moon, ``Active learning for object
  detection with evidential deep learning and hierarchical uncertainty
  aggregation,'' in \emph{The Eleventh International Conference on Learning
  Representations}, 2023.

\bibitem{liu2016ssd}
W.~Liu, D.~Anguelov, D.~Erhan, C.~Szegedy, S.~Reed, C.-Y. Fu, and A.~C. Berg,
  ``Ssd: Single shot multibox detector,'' in \emph{European conference on
  computer vision}.\hskip 1em plus 0.5em minus 0.4em\relax Springer, 2016, pp.
  21--37.

\bibitem{wu2022entropy}
J.~Wu, J.~Chen, and D.~Huang, ``Entropy-based active learning for object
  detection with progressive diversity constraint,'' in \emph{Proceedings of
  the IEEE/CVF Conference on Computer Vision and Pattern Recognition}, 2022.

\bibitem{yu2022consistency}
W.~Yu, S.~Zhu, T.~Yang, and C.~Chen, ``Consistency-based active learning for
  object detection,'' in \emph{Proceedings of the IEEE/CVF Conference on
  Computer Vision and Pattern Recognition}, 2022.

\bibitem{kothawade2022talisman}
S.~Kothawade, S.~Ghosh, S.~Shekhar, Y.~Xiang, and R.~Iyer, ``Talisman: targeted
  active learning for object detection with rare classes and slices using
  submodular mutual information,'' in \emph{European Conference on Computer
  Vision}, 2022.

\bibitem{yuan2023bi3d}
J.~Yuan, B.~Zhang, X.~Yan, T.~Chen, B.~Shi, Y.~Li, and Y.~Qiao, ``Bi3d:
  Bi-domain active learning for cross-domain 3d object detection,'' in
  \emph{Proceedings of the IEEE/CVF Conference on Computer Vision and Pattern
  Recognition}, 2023.

\bibitem{shi20193d}
S.~Shi, X.~Wang, and H.~P. Li, ``3d object proposal generation and detection
  from point cloud,'' in \emph{Proceedings of the IEEE Conference on Computer
  Vision and Pattern Recognition, Long Beach, CA, USA}, 2019, pp. 16--20.

\bibitem{wang2019frustum}
Z.~Wang and K.~Jia, ``Frustum convnet: Sliding frustums to aggregate local
  point-wise features for amodal 3d object detection,'' in \emph{2019 IEEE/RSJ
  International Conference on Intelligent Robots and Systems (IROS)}.\hskip 1em
  plus 0.5em minus 0.4em\relax IEEE, 2019, pp. 1742--1749.

\bibitem{qi2019deep}
C.~R. Qi, O.~Litany, K.~He, and L.~J. Guibas, ``Deep hough voting for 3d object
  detection in point clouds,'' in \emph{Proceedings of the IEEE/CVF
  International Conference on Computer Vision}, 2019, pp. 9277--9286.

\bibitem{qi2017pointnet}
C.~R. Qi, H.~Su, K.~Mo, and L.~J. Guibas, ``Pointnet: Deep learning on point
  sets for 3d classification and segmentation,'' in \emph{Proceedings of the
  IEEE conference on computer vision and pattern recognition}, 2017, pp.
  652--660.

\bibitem{qi2017pointnet++}
C.~R. Qi, L.~Yi, H.~Su, and L.~J. Guibas, ``Pointnet++: Deep hierarchical
  feature learning on point sets in a metric space,'' in \emph{Advances in
  neural information processing systems}, 2017.

\bibitem{wang2019dynamic}
Y.~Wang, Y.~Sun, Z.~Liu, S.~E. Sarma, M.~M. Bronstein, and J.~M. Solomon,
  ``Dynamic graph cnn for learning on point clouds,'' \emph{Acm Transactions On
  Graphics}, 2019.

\bibitem{zhou2018voxelnet}
Y.~Zhou and O.~Tuzel, ``Voxelnet: End-to-end learning for point cloud based 3d
  object detection,'' in \emph{Proceedings of the IEEE conference on computer
  vision and pattern recognition}, 2018, pp. 4490--4499.

\bibitem{chen2020object}
Q.~Chen, L.~Sun, Z.~Wang, K.~Jia, and A.~Yuille, ``Object as hotspots: An
  anchor-free 3d object detection approach via firing of hotspots,'' in
  \emph{European Conference on Computer Vision}.\hskip 1em plus 0.5em minus
  0.4em\relax Springer, 2020, pp. 68--84.

\bibitem{chen2020every}
Q.~Chen, L.~Sun, E.~Cheung, and A.~L. Yuille, ``Every view counts: Cross-view
  consistency in 3d object detection with hybrid-cylindrical-spherical
  voxelization,'' \emph{Advances in Neural Information Processing Systems},
  2020.

\bibitem{li2024unimode}
Z.~Li, X.~Xu, S.~Lim, and H.~Zhao, ``Unimode: Unified monocular 3d object
  detection,'' in \emph{Proceedings of the IEEE/CVF Conference on Computer
  Vision and Pattern Recognition}, 2024, pp. 16\,561--16\,570.

\bibitem{li2024bevnext}
Z.~Li, S.~Lan, J.~M. Alvarez, and Z.~Wu, ``Bevnext: Reviving dense bev
  frameworks for 3d object detection,'' in \emph{Proceedings of the IEEE/CVF
  Conference on Computer Vision and Pattern Recognition}, 2024, pp.
  20\,113--20\,123.

\bibitem{liu2023bevfusion}
Z.~Liu, H.~Tang, A.~Amini, X.~Yang, H.~Mao, D.~L. Rus, and S.~Han, ``Bevfusion:
  Multi-task multi-sensor fusion with unified bird's-eye view representation,''
  in \emph{IEEE International Conference on Robotics and Automation}, 2023.

\bibitem{har2011geometric}
S.~Har-Peled, \emph{Geometric approximation algorithms}.\hskip 1em plus 0.5em
  minus 0.4em\relax American Mathematical Soc., 2011, no. 173.

\bibitem{chen2017multi}
X.~Chen, H.~Ma, J.~Wan, B.~Li, and T.~Xia, ``Multi-view 3d object detection
  network for autonomous driving,'' in \emph{Proceedings of the IEEE Conference
  on Computer Vision and Pattern Recognition}, 2017.

\bibitem{Vora_2020_CVPR}
S.~Vora, A.~H. Lang, B.~Helou, and O.~Beijbom, ``Pointpainting: Sequential
  fusion for 3d object detection,'' in \emph{Proceedings of the IEEE/CVF
  Conference on Computer Vision and Pattern Recognition (CVPR)}, June 2020.

\bibitem{Caesar_2020_CVPR}
H.~Caesar, V.~Bankiti, A.~H. Lang, S.~Vora, V.~E. Liong, Q.~Xu, A.~Krishnan,
  Y.~Pan, G.~Baldan, and O.~Beijbom, ``nuscenes: A multimodal dataset for
  autonomous driving,'' in \emph{Proceedings of the IEEE/CVF Conference on
  Computer Vision and Pattern Recognition (CVPR)}, June 2020.

\bibitem{ash2019deep}
J.~T. Ash, C.~Zhang, A.~Krishnamurthy, J.~Langford, and A.~Agarwal, ``Deep
  batch active learning by diverse, uncertain gradient lower bounds,'' in
  \emph{International Conference on Learning Representations}, 2020.

\bibitem{yang2024ppal}
{Yang, Chenhongyi and Huang, Lichao and Crowley, Elliot J.}, ``{Plug and Play
  Active Learning for Object Detection},'' in \emph{Proceedings of the IEEE/CVF
  Conference on Computer Vision and Pattern Recognition (CVPR)}, 2024.

\bibitem{uwe2024}
Y.~He, L.~Cai, J.~Liao, and C.-S. Foo, ``Hybrid active learning with
  uncertainty-weighted embeddings,'' \emph{Transactions on Machine Learning
  Research}, 2024.

\bibitem{holub2008entropy}
A.~Holub, P.~Perona, and M.~C. Burl, ``Entropy-based active learning for object
  recognition,'' in \emph{2008 IEEE Computer Society Conference on Computer
  Vision and Pattern Recognition Workshops}.\hskip 1em plus 0.5em minus
  0.4em\relax IEEE, 2008.

\bibitem{yuan2020cold}
M.~Yuan, H.-T. Lin, and J.~Boyd-Graber, ``Cold-start active learning through
  self-supervised language modeling,'' \emph{arXiv preprint arXiv:2010.09535},
  2020.

\end{thebibliography}
%\end{IEEEbiography}
%% insert where needed to balance the two columns on the last page with
%%\newpage


%\vfill


\end{document}